\documentclass{article}

\usepackage[final]{neurips_2024}

\usepackage{amsmath,amsfonts}
\usepackage{array}
\usepackage{textcomp}
\usepackage{url}
\usepackage{graphicx}
\usepackage{subcaption}

\usepackage{booktabs}  
\usepackage{makecell}
\usepackage{paralist}
\usepackage{multirow}
\usepackage{float}
\usepackage{tabularx}
\usepackage{xcolor}
\usepackage{hyperref}
\usepackage{orcidlink}
\usepackage{pifont}

\newcommand{\ra}[1]{\renewcommand{\arraystretch}{#1}}
\newcommand{\best}[1]{\textcolor{black}{\textbf{#1}}}
\newcommand{\secondbest}[1]{\textcolor{black}{#1}}
\newcommand{\cmark}{\ding{51}}
\newcommand{\xmark}{\ding{55}}
\newcommand{\lstm}{MB-ConvLSTM}
\newcommand{\sys}{DeepLight}

\newcommand{\X}{\mathcal{X}}

\newcommand{\Hc}{\mathcal{H}}
\newcommand{\Xb}{\overline{\mathcal{X}}}
\newcommand{\Hb}{\overline{\mathcal{H}}}
\newcommand{\Hh}{\mathcal{H}}
\newcommand{\C}{\mathcal{C}}

\usepackage{enumitem}
\usepackage[T1]{fontenc}

\def\tsc#1{\csdef{#1}{\textsc{\lowercase{#1}}\xspace}}
\tsc{WGM}
\tsc{QE}
\tsc{EP}
\tsc{PMS}
\tsc{BEC}
\tsc{DE}

\title{Lightning Prediction under Uncertainty: DeepLight with Hazy Loss} 

\author{%
  Md Sultanul Arifin\textsuperscript{1}
    \\
  \texttt{1805097@ugrad.cse.buet.ac.bd} \\
  \And
  Abu Nowshed Sakib\textsuperscript{1} \\
  \texttt{1705107@ugrad.cse.buet.ac.bd} \\
  \AND
  Yeasir Rayhan\textsuperscript{2} \\
  \texttt{yrayhan@purdue.edu} \\
  \And
  Tanzima Hashem\textsuperscript{1} \\
  \texttt{tanzimahashem@cse.buet.ac.bd} \\
}

\begin{document}
\let\WriteBookmarks\relax
\def\floatpagepagefraction{1}
\def\textpagefraction{.001}
\maketitle

\begin{center}
\textsuperscript{1} Department of Computer Science and Engineering \\
Bangladesh University of Engineering and Technology \\
  Dhaka 1000, Bangladesh \\

\vspace{0.8em}
  
\textsuperscript{2}  Purdue University \\
   West Lafayette, IN, USA \\
   
\end{center}

\begin{abstract}
Lightning, a common feature of severe meteorological conditions, poses significant risks, from direct human injuries to substantial economic losses. These risks are further exacerbated by climate change. Early and accurate prediction of lightning would enable preventive measures to safeguard people, protect property, and minimize economic losses. In this paper, we present DeepLight, a novel deep learning architecture for predicting lightning occurrences. Existing prediction models face several critical limitations: i) they often struggle to capture the dynamic spatial context and the inherent randomness of lightning events, including whether lightning occurs and its variability in location and timing even under similar meteorological conditions; ii) they underutilize key observational data, such as radar reflectivity and cloud properties; and iii) they rely heavily on Numerical Weather Prediction (NWP) systems, which are both computationally expensive and highly sensitive to parameter settings. To overcome these challenges, DeepLight leverages multi-source meteorological data, including radar reflectivity, cloud properties, and historical lightning occurrences through a dual-encoder architecture. By employing multi-branch convolution techniques, it dynamically captures spatial correlations across varying extents. Furthermore, its novel Hazy Loss function explicitly addresses the spatio-temporal uncertainty of lightning by penalizing deviations based on proximity to true events, enabling the model to better learn patterns amidst randomness. Extensive experiments show that DeepLight improves the Equitable Threat Score (ETS) by 18\%--30\% over state-of-the-art methods, establishing it as a robust solution for lightning prediction.
\end{abstract}





\section{Introduction}
\label{sec:introduction}

Lightning, a typical characteristic of severe meteorological conditions, poses significant risks including fatalities, injuries, property damage, and disruptions to electronic and aviation systems~\citep{cooper2019reducing}. For example, in Bangladesh, 43 people lost their lives due to lightning strikes in just eight days (May 1--8, 2024), a trend linked to climate change\footnote{https://perma.cc/EPH5-9289}. Similarly, Nepal recorded 360 lightning-related deaths between 2019 and 2023, surpassing annual monsoon flood fatalities\footnote{https://perma.cc/58YD-YSEN}. Beyond direct casualties, lightning triggers wildfires and disrupts critical infrastructure, such as the 2012 lightning-induced wildfires in the United States of America, which burned over 9 million acres~\citep{cooper2019reducing}. These impacts underscore the importance of accurate early lightning prediction.

Predicting lightning is challenging due to its highly localized, transient, and inherently random nature, whereby lightning may or may not occur, or may occur at different locations and times, even under similar observed meteorological conditions. This form of uncertainty is distinct from epistemic uncertainty or uncertainty characterized through probability distributions, variance, or confidence intervals. Lightning forms within rapidly evolving convective storms where microphysical interactions among ice, graupel, and supercooled water lead to charge separation. These electrification processes occur on millisecond scales and are not directly observable by standard atmospheric sensors, limiting the predictability of lightning occurrence. As a result, researchers rely on historical lightning records and indirect indicators such as radar reflectivity and cloud properties~\citep{raheem2023techniques} to predict lightning.

Earlier works on lightning prediction have relied heavily on Numerical Weather Prediction (NWP, for short) systems~\citep{powers2017weather}. Empirical schemes such as PR92~\citep{price1992simple} and MNSRP99~\citep{Michalon1999} use NWP-simulated microphysical and dynamical parameters to estimate lightning frequency. However, NWP-based methods are sensitive to physics parameterization choices, require extensive computational resources, and struggle to capture fine-scale spatial patterns.

Recent deep learning approaches (e.g., LightNet~\citep{lightnet}, ADSNet~\citep{adsnet}, HSTN~\citep{HSTN}, LightNet+~\citep{zhou2022lightnet+}) have improved lightning prediction by learning from historical lightning observations and NWP-based features. Yet, these models face key limitations: (i) difficulty in modeling dynamic spatial contexts of varying extents, (ii) limited integration of real observational radar and cloud data, and (iii) continued reliance on NWP systems despite their known variability in outcomes based on different physics parameterization schemes for the same storm event~\citep{wes-7-1869-2022}. These limitations highlight the need for an NWP-independent approach that can robustly learn spatiotemporal correlations from real multi-source data.

In this paper, DeepLight is developed as a novel deep learning–based framework for lightning prediction. DeepLight learns the spatiotemporal correlations between meteorological parameters and lightning occurrences, as well as the interdependencies among these parameters, using only real observational data. DeepLight introduces a multi-branch architecture capable of modeling spatial dependencies across varying scales. It incorporates the Hazy Loss function, which addresses the inherent randomness and uncertainty in lightning events by penalizing spatio-temporal deviations from the ground truth based on their proximity. This encourages the model to learn patterns that can tolerate variability in spatial and temporal occurrence. By integrating radar reflectivity, cloud properties, and historical lightning observations, DeepLight provides a comprehensive and robust solution to the problem of lightning prediction. Our model design eliminates the reliance on NWP systems while addressing the dynamic spatial context and the inherent randomness or irregularity in lightning occurrences.

Our contributions are summarized as follows:
\begin{itemize}
    \item We propose DeepLight, a deep learning architecture for lightning prediction that leverages real multi-source data and removes dependency on NWP systems.
    \item We identify the dynamic nature of spatial correlation extents and design multi-branch convolution techniques to capture context from varying spatial ranges.
    \item We introduce Hazy Loss, a neighborhood-aware loss function that improves learning under high spatiotemporal uncertainty that is inherent in lightning events.
    \item We demonstrate through extensive experiments that DeepLight substantially outperforms existing models under multiple forecast horizons.
\end{itemize}

\section{Problem Formulation}
\label{sec:prob}
We aim to predict future lightning occurrences using historical real-world data, including lightning observations and activities, and auxiliary meteorological parameters such as radar reflectivity~\citep{noaa_radar_reflectivity} and cloud properties~\citep{abi-cod, abi-acha, abi-ctp}. The selection of these features is guided by the comprehensive analysis by Leinonen et al.~\citep{nhess-22-577-2022}, which underscores their importance in lightning forecasting. Additionally, previous state of the art deep learning studies~\citep{lightnet, adsnet, zhou2022lightnet+} have leveraged these parameters, either in simulated or real-world formats, to improve prediction performance. We assume that the target region is divided into an $N\times N$ grid, where each grid cell represents a spatial unit for which lightning occurrence is predicted. We next define lightning and meteorological parameters used in lightning prediction:

\vspace{4pt}
\noindent \emph{Lightning Occurrence} (\boldmath{$L_t$}): Lightning occurrence (\boldmath{$L_t$}) for a region denotes whether lightning occurs or not at $t$-th time-step: $[t, t+1)$.

\vspace{4pt}
\noindent \emph{Lightning Activities} (\boldmath{$A_t$}): Lightning activities (\boldmath{$A_t$}) is represented as $A_t$ = [Flash Frequency, Flash Energy]. Specifically, flash frequency quantifies the number of occurred lightning flashes and flash energy measures the total energy released by lightning flashes of a region at $t^{th}$ time-step: $[t, t+1)$.

\vspace{4pt}
\noindent \emph{Radar Reflectivity} (\boldmath{$R_t$}): In thunderstorm clouds, charge develops through the Triboelectric Effect, caused by friction between differently sized hydrometeors~\citep{illingworth1985charge}. Hydrometeors of various shapes and sizes react differently to radio waves. Radar reflectivity values \boldmath{$R_t$} provide information about these hydrometeors at $t^{th}$ time-step: $[t, t+1)$, indicating the potential for charge buildup and lightning occurrences.

\vspace{4pt}
\noindent \emph{Cloud Properties} (\boldmath{$D_t$}): Cloud behaves like a giant capacitor where the upper (lighter) portion of the cloud is positively charged and lower (heavier) portion is negatively charged, storing electrical energy until it is discharged as lightning. Cloud Properties are represented as 
$D_t$ = [Cloud Top Height, Cloud Top Pressure, Cloud Optical Depth] for the $t^{th}$ time-step: $[t, t+1)$.
\begin{compactitem}
    \item 
    Cloud top height signifies the geopotential height at the top of a cloud layer, measured in feet.    
    \item 
    Cloud top pressure denotes the pressure reading at the top of a cloud layer, measured in hectopascals (hPa).
    \item 
    Cloud optical depth refers to the vertical optical thickness of the cloud, determined by particle composition, form, concentration, and extent.    
\end{compactitem}

\begin{figure}[htbp]
\centering
  \includegraphics[width=0.8\textwidth]{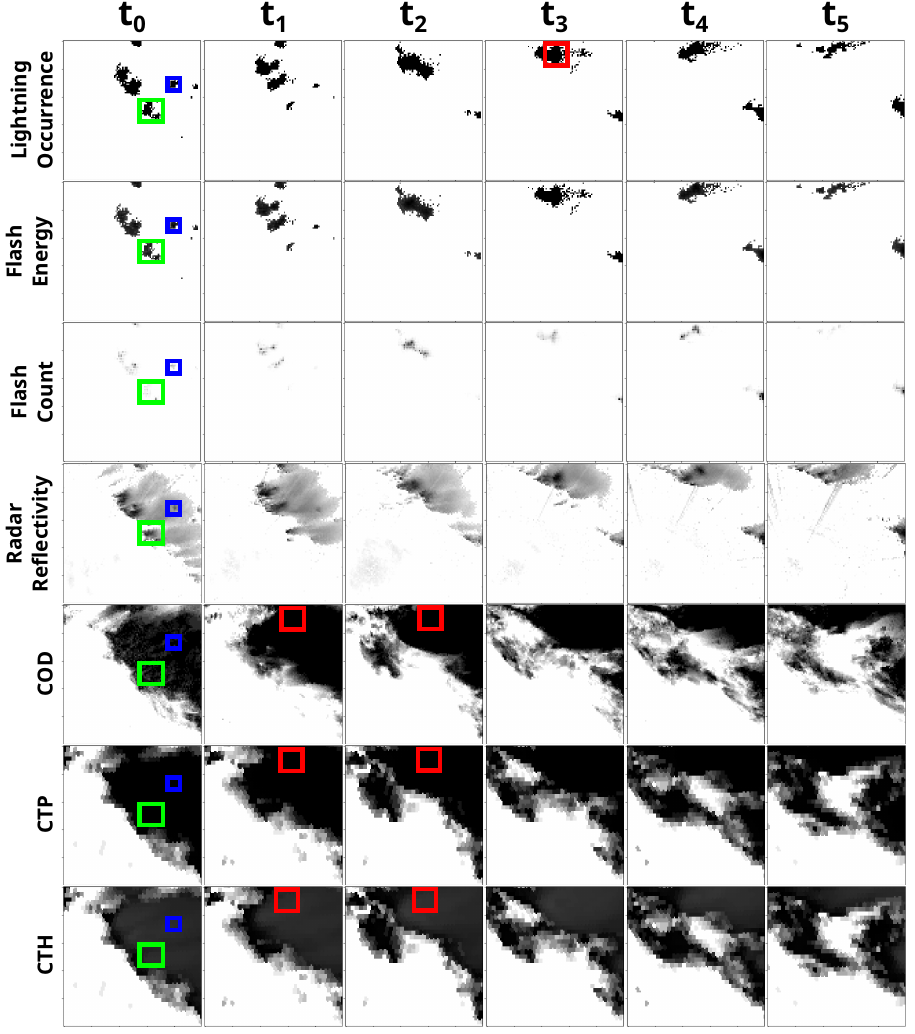}
  \caption{Temporal and spatial correlations among lightning and other meteorological parameters. (\textit{Note:} COD = Cloud Optical Depth; CTP = Cloud Top Pressure; CTH = Cloud Top Height from AWG Cloud Height Algorithm. Details of these parameters are discussed in Section~\ref{sec:prob}.)}
  \label{fig:storm}
\end{figure}

\vspace{1mm}
\noindent\textbf{Problem Definition:}
Given lightning observations ($L_t$), lightning activities ($A_t$), radar reflectivity ($R_t$) and cloud properties ($D_t$) for the last $s$ time-steps (i.e., $t = -s, \ldots, -2, -1$) for each cell of an $N \times N$ grid, the objective is to forecast the probabilistic estimate ($\hat{L}_t$; $\hat{L}_t \in [0, 1]$) denoting the likelihood of lightning occurrence for the future $h$ time steps (i.e., $t = 0, 1, 2, \ldots, h-1$) for each cell of the grid. For evaluation and decision-making, this probabilistic output is converted into a binary lightning/no-lightning prediction using a predefined probability threshold.
\vspace{1mm}

\emph{Correlations.} Lightning and other meteorological parameters exhibit both visible and latent temporal and spatial correlations with lightning occurrences (Figure~\ref{fig:storm}), and recognizing these patterns is crucial for accurate lightning prediction. For instance, cloud presence is observed in the red region at timestamps $t_1$ and $t_2$ (COD, CTP, and CTH), preceding the occurrence of lightning in the same region at $t_3$ (1st Row), illustrating temporal correlation. Conversely, at the same timestamp $t_0$, higher values across multiple parameters are observed in the green region, indicating spatial correlation with lightning flashes occurring in that region. Moreover, the spatial extent over which such correlations manifest is dynamic. As shown in Figure~\ref{fig:storm}, the green and blue regions differ in size but exhibit similar spatial correlation patterns. We identify this notion of dynamic spatial extent and incorporate it into our model, enabling more adaptive and accurate lightning prediction.

\section{Related Works}
\label{sec:related}
\subsection{Lightning Prediction Models}
\label{sec:lightning}

Traditionally, Numerical Weather Prediction (NWP) systems have been used to forecast lightning occurrences. One of such popular NWP systems is the Weather Research and Forecasting (WRF) model~\citep{powers2017weather}. WRF is a mesoscale NWP system designed for atmospheric research and operational forecasting, capable of simulating a wide range of weather phenomena. Recent research has attempted to solve the lightning prediction problem in various ways. Some models incorporate machine learning techniques while relying on features derived from NWP-based simulations or real observational data. Others introduce novel loss functions or specialized deep learning modules to enhance predictive accuracy. Table~\ref{tab:related_work} gives an overview of the existing work on lightning prediction and how they differ from our proposed model in terms of correlation modeling, methodologies and used features.

\subsubsection{Correlation Modeling} Temporal and spatial correlations are evident in lightning occurrences with varying spatial extent. Numerical Weather Prediction (NWP) systems, such as~\citep{powers2017weather} primarily rely on physics parameterization schemes to capture temporal correlations and simulate various atmospheric parameters, e.g., max vertical velocity and precipitation. Empirical methods like PR92~\citep{price1992simple} and MNSRP99~\citep{Michalon1999} use these simulated parameters to predict lightning. However, previous studies~\citep{lightnet, adsnet, zhou2022lightnet+} have shown that these methods have inherent limitations when it comes to capturing spatial correlations. Deep learning based approaches such as LightNet~\citep{lightnet}, ADSNet~\citep{adsnet}, HSTN~\citep{HSTN} and LightNet+~\citep{zhou2022lightnet+}, can model both spatial and temporal patterns in data upto a certain level. However, these models still struggle to effectively handle different spatial extent of the correlations. DeepLight significantly improves lightning prediction accuracy by effectively capturing temporal correlations and spatial correlations of varying extent.

\begin{normalsize}
\setlength\heavyrulewidth{0.20ex}
\begin{table*}[t]
\ra{1.7}
\caption{Comparison of related studies. 
\textmd{
\textbf{S}: Simulated, \textbf{LO}: Lightning Observation, \textbf{CP}: Cloud Properties, \textbf{RR}: Radar Reflectivity, \textbf{WBCE}: Weighted Binary Cross Entropy Loss, \textbf{MSPL}: Multi-scale Pooling Loss, \textbf{GD}: Gaussian Diffusion Module}}
\resizebox{\textwidth}{!}{
\begin{tabular}{@{}lccccccccccc@{}} \toprule

\multirow{2}{*}{\large Study} & \multicolumn{3}{c}{\large Correlation Modelling} & \phantom{abc} & \multicolumn{2}{c}{\large Methodologies} & \phantom{abc} & \multicolumn{4}{c}{\large Features} \\

\cmidrule{2-4} \cmidrule{6-7} \cmidrule{9-12}
& {\large Temporal} & \makecell{\large Spatial} & \makecell{\large Spatial extent} && {\large Loss func.} & {\large Approach} && {\large S} & \makecell{\large LO} & \makecell{\large CP} & \makecell{\large RR} \\
\midrule
PR92\cite{price1992simple} & \cmark & \xmark & \xmark && \xmark & NWP && \cmark & \xmark & \xmark & \xmark \\
MNSRP99\cite{Michalon1999} & \cmark & \xmark & \xmark && \xmark & NWP && \cmark & \xmark & \cmark & \xmark \\
LightNet\cite{lightnet} & \cmark & \cmark & static && \makecell{WBCE} & \makecell{NWP+CLSTM} && \cmark & \cmark & \xmark & \cmark \\
ADSNet\cite{adsnet} & \cmark & \cmark & static && \makecell{WBCE} & \makecell{NWP+Attention+CLSTM} && \cmark & \cmark & \xmark & \xmark \\
HSTN\cite{HSTN} & \cmark & \cmark & static && \makecell{MSPL} & \makecell{NWP+GD+CLSTM} && \cmark & \cmark & \xmark & \cmark \\
LightNet+\cite{zhou2022lightnet+} & \cmark & \cmark & static && \makecell{WBCE} & \makecell{NWP+Attention+CLSTM} && \cmark & \cmark & \xmark & \xmark  \\
\sys & \cmark & \cmark & dynamic && \makecell{Hazy Loss} & \makecell{MB-ConvLSTM} && \xmark & \cmark & \cmark & \cmark  \\
\toprule
\end{tabular}
}
\label{tab:related_work}
\end{table*}
\end{normalsize}

\subsubsection{Methodologies} Approaches in lightning prediction have evolved from traditional statistical methods to deep learning algorithms. Price and Rind~\citep{price1992simple} established a relationship between lightning frequency and maximum vertical velocity, introducing the PR92 lightning parameterization scheme. Later, Michalon et al.~\citep{Michalon1999} proposed that lightning frequency can be represented by a power function of both the cloud top height and the cloud droplet concentration, thereby partially acknowledging the influence of micro-physical cloud properties on lightning occurrences. 

Prediction capabilities of such methods that solely rely on NWP systems are hampered by their inability to calibrate to the observed historical data. This problem is tackled by the deep learning methods~\citep{lightnet, adsnet, HSTN, zhou2022lightnet+} using  hybrid neural network architecture alongside NWP systems to learn from historical lightning occurrence data. Motivated by these limitations, our work, DeepLight, operates without reliance on NWP systems and instead focuses on directly modeling spatiotemporal context from observational data. There exists another study~\citep{cintineo2022probsevere} that does not use NWP systems, rather applies a basic UNet architecture~\citep{ronneberger2015u} for lightning forecasting.

Recent studies in the spatio-temporal prediction field have introduced specialized loss functions beyond the traditional Weighted Binary Cross Entropy (WBCE) loss to enhance model performance. For example, HSTN~\citep{HSTN} proposed Multi-Scale Pooling Loss, which effectively incorporates proximity to the ground truth in the loss computation. These developments motivate us to design a loss function that explicitly and smoothly incorporate spatiotemporal proximity into the learning objective.

\subsubsection{Features} 
Leinonen et al.~\citep{nhess-22-577-2022} conducts an extensive study on the effects of various features on lightning, analyzing the impact of 106 different prediction variables. PR92~\citep{price1992simple} utilizes simulated maximum vertical velocity as a key feature to establish a correlation with lightning frequency. Michalon et al.~\citep{Michalon1999} incorporates simulated micro-physical cloud properties, such as cloud top height and cloud droplet concentration, into their model. In a more comprehensive approach, LightNet~\citep{lightnet} integrates a suite of simulated micro-physical parameters, i.e., ice, snow and graupel mixing ratios, simulated radar reflectivity and maximum vertical velocity derived from the WRF model, and real-world lightning observations. Both ADSNet~\citep{adsnet} and LightNet+~\citep{zhou2022lightnet+} follow a similar feature set to LightNet, with the notable substitution of radar reflectivity with precipitation. HSTN~\citep{HSTN} also uses three observational data points from weather stations: average temperature, average relative humidity, and precipitation. Since the quality of simulated data is sensitive to parameter settings, DeepLight exploits only real-world lightning observations, cloud properties, radar reflectivity for lightning prediction.

\subsection{Spatiotemporal Prediction Models}
\label{sec:spatio-temporal}

Spatiotemporal prediction is central to a wide range of applications, including traffic~\citep{li2022deep, liu2023spatio, zhang2018long}, mobility~\citep{zhang2017deep, zhang2018predicting, tang2022sprnn}, accident~\citep{bao2020uncertainty}, crime~\citep{chen2023spatio}, and air quality forecasting~\citep{chen2024spatiotemporal, huang2021spatio}. These domains typically rely on deep learning architectures to model spatial and temporal dependencies. 

Foundational models like convolutional neural networks (CNNs)~\citep{lecun1998gradient} and recurrent neural networks (RNNs)~\citep{elman1990finding} provided early progress in spatial and temporal modeling, respectively. Long Short-Term Memory (LSTM)~\citep{Hochreiter} networks extended the temporal depth of RNNs, and ConvLSTM~\citep{convlstm} architectures attempted to merge spatial and temporal reasoning. CNN-based models have been successfully applied to event recognition tasks involving high-frequency temporal signals, such as lightning electric field waveform classification, demonstrating their effectiveness in capturing localized spatiotemporal patterns after appropriate preprocessing and optimization~\citep{wang2025application}. Related CNN-based approaches have also been explored in computer vision tasks such as illumination estimation, where global lighting conditions are inferred from spatial image cues~\citep{buyukarikan2022using}. However, these methods face challenges in capturing complex dependencies and uncertainty in many real-world tasks.

Recent advances have addressed limitations of early CNN and RNN-based models by developing architectures tailored for complex spatiotemporal dependencies and uncertainty. Self-supervised methods like SelfWeather~\citep{gong2024spatio} leverage contrastive and generative objectives for robust feature learning without heavy labeling. StepDeep~\citep{stepdeep} employs 3D convolutions for dense spatial-temporal encoding but can struggle with stochastic phenomena such as lightning. Spiking neural networks (SNNs) combined with spatial-temporal self-attention (STS-Transformer)~\citep{wang2023spatial} provide asynchronous, energy-efficient modeling with enhanced relative position bias.

In weather forecasting, GraphCast~\citep{lam2023learning} uses graph neural networks trained on reanalysis data to outperform traditional deterministic models, while Pangu-Weather~\citep{bi2022pangu, bi2023accurate} integrates 3D Earth-specific transformers and hierarchical temporal aggregation, achieving strong generalization and superior cyclone tracking through extensive historical data training. For renewable energy, HSTTN~\citep{zhang2023long} introduces an hourglass-shaped Transformer network with skip connections and contextual fusion to jointly model hierarchical temporal scales and spatial correlations, excelling in long-term wind power forecasting. In urban systems, DMVST-Net~\citep{yao2018deep} combines LSTM, local CNN, and semantic views in a multi-view spatiotemporal framework, significantly improving taxi demand prediction by capturing complex nonlinear dependencies across space and time.

While these spatiotemporal models have demonstrated strong performance in their respective domains, they are not directly suited to lightning prediction. Lightning events are highly localized, transient, and inherently uncertain, with longer dynamic spatiotemporal correlations than those in large-scale traffic, mobility, or weather systems. Moreover, most existing models do not explicitly address the stochastic nature of lightning formation, nor do they leverage the unique combination of real-world radar reflectivity, cloud properties, and historical lightning observations without reliance on computationally expensive Numerical Weather Prediction systems. Consequently, specialized architectures and loss functions are necessary to effectively capture the dynamic spatial extent and the high spatio-temporal uncertainty inherent in lightning forecasting.

\section{Materials and Methods}
\label{sec:model}
In this paper, we introduce \emph{DeepLight}, a 
deep learning model for predicting lightning occurrences. The model's improved performance is driven by its novel neighborhood-aware loss function called \emph{Hazy Loss}, its multi-branch deep learning architecture, and its ability to learn from diverse, real-world meteorological observation data, including radar reflectivity, cloud properties, and historical lightning occurrences. The \emph{Hazy Loss} function 
applies 
smooth scoring to manage the randomness of lightning events, penalizing spatio-temporal deviations to address key forecasting challenges effectively. This penalization scheme helps model predict more closely to the region of actual lightning occurrence. The multi-branch deep learning architecture enables the model to adaptively capture the dynamic spatial extent of meteorological phenomena. For instance, as illustrated in Figure~\ref{fig:storm}, lightning occurrence patterns often manifest in clustered formations of varying sizes (e.g., the large green box and small blue box at $t_0$). The multi-branch approach allows the model to assign different kernel sizes to the horizontally stacked convolution layers, effectively adjusting the field of focus to accommodate variations in spatial extent, thereby enhancing its ability to capture patterns of different sizes.

\begin{figure}[htbp]
    \centering   
    \includegraphics[width=\textwidth]{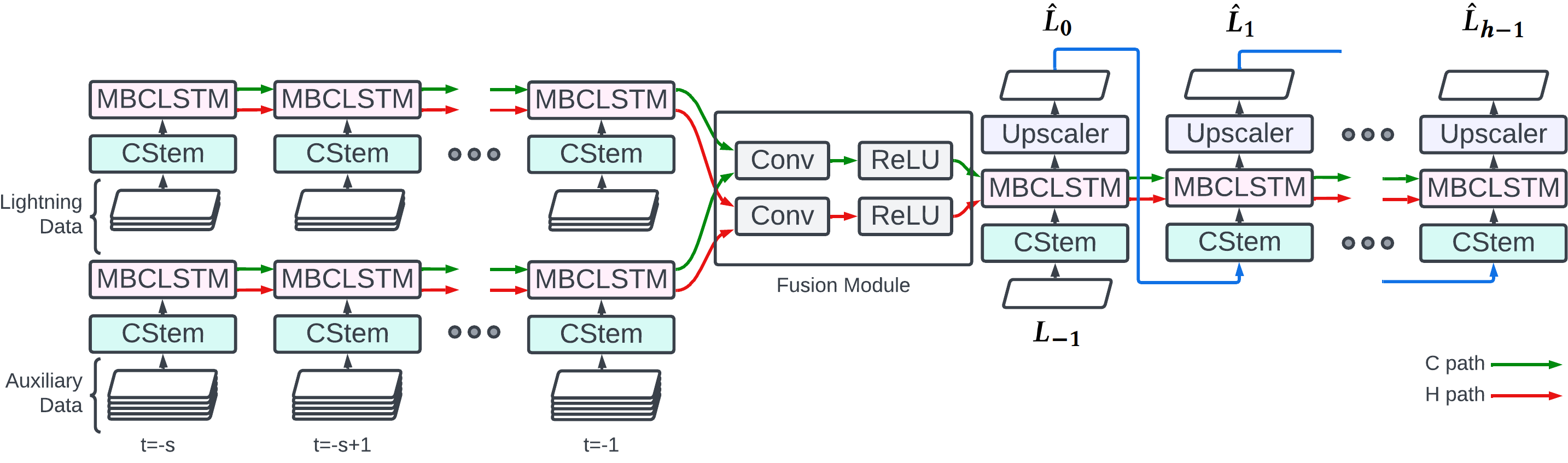}
    \caption{Network Architecture of \sys{}. 
    (\textit{Note:} MBCLSTM = Multi Branch Convolutional LSTM; CStem = Convolutional Stem)}
    \label{fig:model}
\end{figure}

\sys{} adopts a dual encoder-decoder architecture consisting of two encoders and a single decoder (cf. Figure~\ref{fig:model}). The two encoders model the lightning data, e.g., lightning occurrence, lightning flash count, intensity, and the accompanying meteorological condition data, e.g., Cloud Properties, Radar Reflectivity etc., respectively. Each encoder consists of a single convolutional stem (CStem) (cf. Section~\ref{subsubsec:cstem}) followed by a Multi-Branch ConvLSTM (\lstm{}) (cf. Section~\ref{subsubsec:mbclstm}), and generates two separate contexts representing the lightning observation and the accompanying meteorological condition. These two contexts are then fused together and fed to the decoder. The decoder consists of a convolution stem (CStem)  followed by a Multi-Branch ConvLSTM (\lstm{}), the output of which is upscaled by Transposed Convolution~\citep{zeiler2010deconvolutional}.

\subsection{Multi-Branched Approach}

Multi-branching~\citep{inception} is a deep learning design paradigm that employs multiple parallel pathways within a single module, allowing the network to process inputs through diverse transformations. This concept is broadly used in various architectures, including multi-head attention in transformers and multi-path convolutional networks in computer vision. In the context of lightning prediction, capturing spatial correlations at varying scales is crucial due to the dynamic nature of lightning clusters. To address this, our model leverages multi-branching within both CStem and \lstm{}.

\begin{figure*}[t]
    \begin{subfigure}[b]{0.40\textwidth}
        \centering
        \includegraphics[height=2.2in]{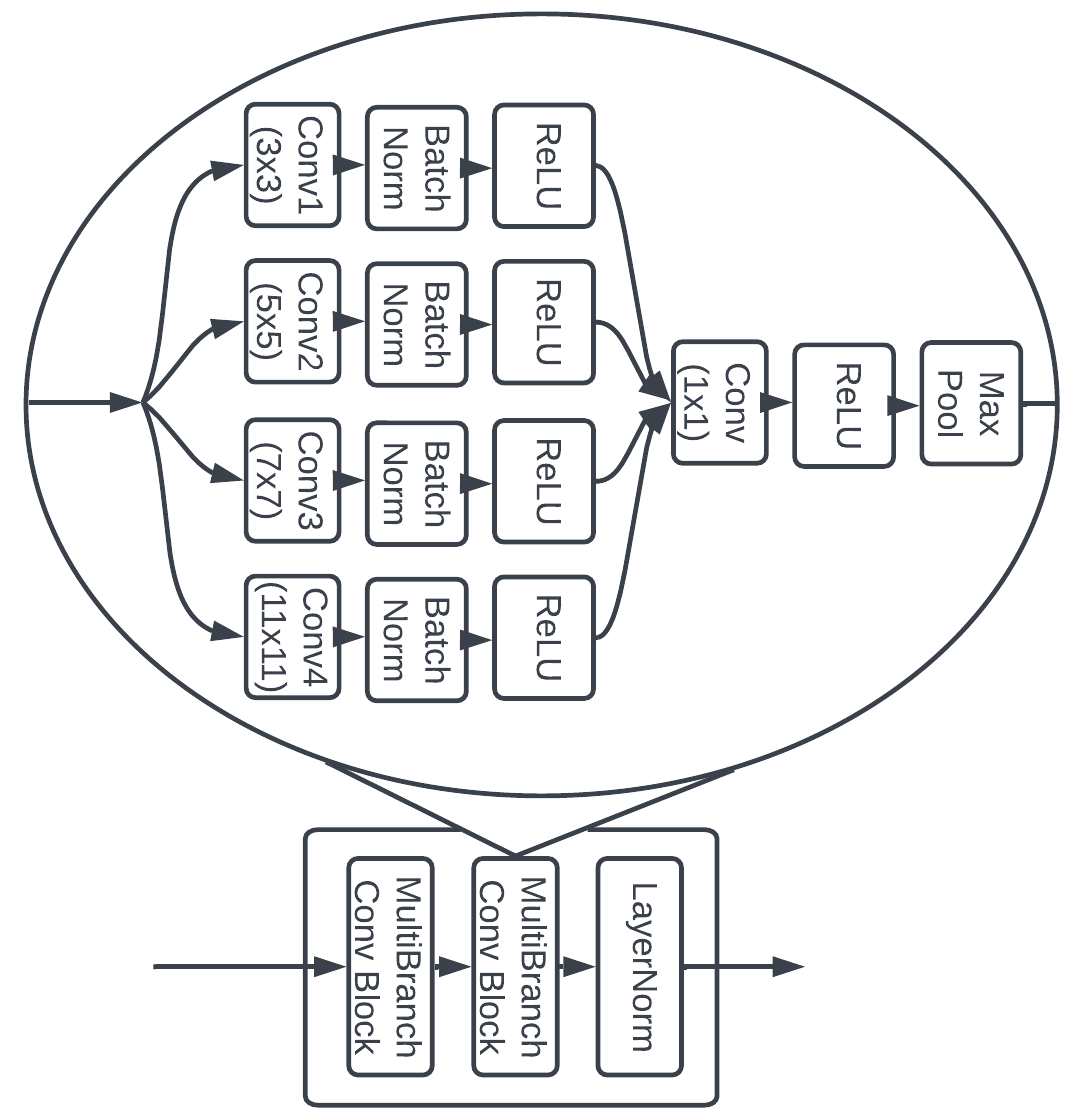}
        \caption{Convolutional Stem (CStem).}
        \label{fig:convstem_multipath}
    \end{subfigure}%
    \begin{subfigure}[b]{0.47\textwidth}
        \centering
        \includegraphics[height=2.2in]{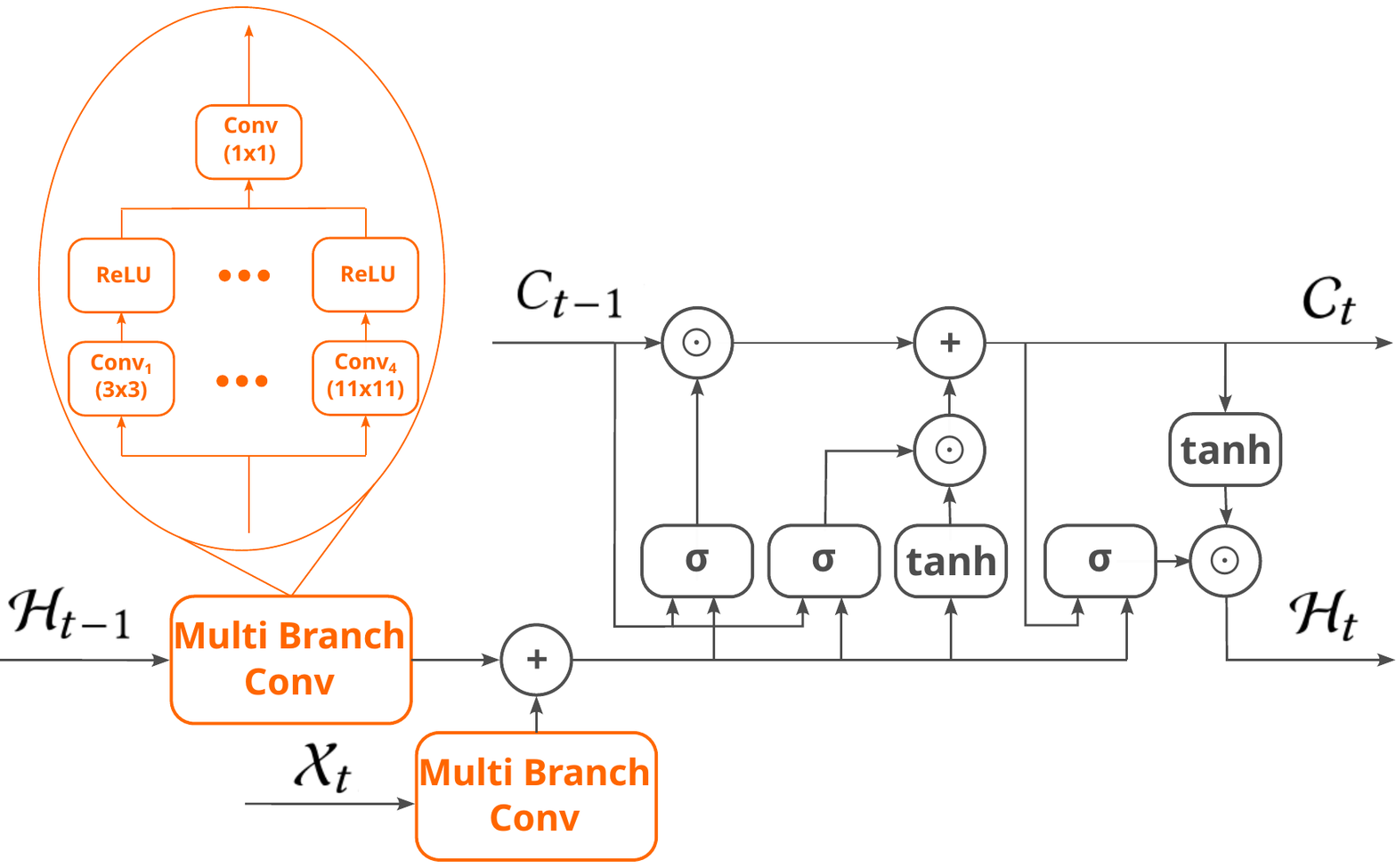}
        \caption{Multi-Branch ConvLSTM.}
        \label{fig:modified_convlstm}
    \end{subfigure}
    \caption{Multi-Branched approaches used in \sys{}}
\end{figure*}

\subsubsection{Convolutional Stem (CStem)}
\label{subsubsec:cstem}

A convolutional stem (CStem) downsamples the input through a series of convolution operations. CStem of previous studies~\citep{lightnet, adsnet} consists of multiple  $p\times p$ fixed sized convolutions layers stacked vertically (one after another). In case of lightning, this imposes a hard restriction on the radius of the influence of nearby cells on a target cell. This results in the model learning incomplete information, as lightning storms can cover distances depending on the strength of the storm. Hence, we modify the CStem and introduce MultiBranch Conv Block in it (cf. Figure~\ref{fig:convstem_multipath}). This block stacks multiple $p\times p$ convolutions of different size horizontally (side-by-side) and let the network learn what convolutions to focus more. This improves the generalization capability of the network, and helps the model learn better lightning representation. 

Note that, even though the underlying motivation of ours matches with the Inception module of GoogleNet~\citep{inception}, there is a difference in the implementation details. The Inception module employs multiple convolutional filters of different sizes (e.g., \(1 \times 1\), \(3 \times 3\), \(5 \times 5\)) in parallel to capture features at various scales, followed by concatenation of their outputs. It also includes a dimensionality reduction step using \(1 \times 1\) convolutions before the larger filters to reduce computational cost. In contrast, our multi-branch convolution block directly stacks multiple \(p \times p\) convolutions of different sizes side by side without preliminary dimension reduction. Each branch is followed by batch normalization and ReLU activation, and the outputs are concatenated and fused using a final \(1 \times 1\) convolution, followed by max pooling. This design allows our model to dynamically learn the relevant spatial extent without enforcing architectural constraints like pre-activation dimensionality reduction, making it better suited for capturing variable-scale patterns in meteorological data.

\subsubsection{Multi Branch Convolutional LSTM (\lstm{})}
\label{subsubsec:mbclstm}

MB-ConvLSTM is built upon the standard ConvLSTM architecture but addresses a critical limitation: ConvLSTM employs a fixed-sized $p\times p$ convolution, which restricts its ability to capture dynamic spatial dependencies. As discussed, this limitation is particularly problematic for lightning prediction, where lightning clusters vary significantly in size and shape due to complex atmospheric interactions. ConvLSTM’s fixed receptive field prevents it from effectively adapting to these variations, leading to suboptimal performance in capturing meteorological patterns of varying sizes.

To overcome this, we propose MB-ConvLSTM (Figure~\ref{fig:modified_convlstm}), which introduces two separate but identical Multi-Branch Convolution blocks applied to \textbf{Hidden state} and \textbf{Input}, respectively to relax the rigid locality constraint of ConvLSTM. The Multi-Branch Convolution block used in MB-ConvLSTM closely resembles the one employed in our Convolutional Stem (CStem) as described in Section~\ref{subsubsec:cstem}, with a few key differences: (i) batch normalization is omitted within each branch, and (ii) the fused output from the final \(1 \times 1\) convolution is not followed by ReLU activation or max pooling. These modifications help preserve the temporal dynamics within the recurrent unit while still enabling the model to capture spatial features across varying receptive fields.

The driving equations of \lstm{} are presented as follows, where $\circ$ and $*$ represent the Hadamard product and convolution operation, respectively. $K_p$ represents a $p\times p$ convolution, and $\mathbin\Vert$ represents the concatenation operation. Let $\X_t$, $\Hc_t$, $\C_t$ be the input, hidden state and cell state, of \lstm{}, respectively. Let $i_t$, $f_t$, and $o_t$ be the input, forget and output gates of the \lstm{}, respectively.
\begin{align}
\Xb_t &= K_{1} *( \mathbin\Vert_{p\in \{3,5,7,11\}} ( ReLU (K_p * \X_t)))\\
\Hb_{t-1} &= K_{1} *( \mathbin\Vert_{p\in \{3,5,7,11\}} ( ReLU (K_p * \Hh_{t-1})))\\
f_t &= \sigma( \left[\Xb_t\right]^1 + \left[\Hb_{t-1}\right]^1 +  W_{cf} \circ \C_{t-1} + b_f) \\
i_t &= \sigma(\left[\Xb_t\right]^2 + \left[\Hb_{t-1}\right]^2 + W_{ci} \circ \C_{t-1} + b_i) \\
\C_t &= f_t \circ \C_{t-1} + i_t \circ \tanh( \left[\Xb_t\right]^3 + \left[\Hb_{t-1}\right]^3 + b_c) \\
o_t &= \sigma(\left[\Xb_t\right]^4 + \left[\Hb_{t-1}\right]^4 + W_{co} \circ \C_{t} + b_o ) \\
\Hc_t &= o_t \circ \tanh(\C_t)
\end{align}  

Here, to determine the future state of a certain cell, the input $\X_t$ and the hidden state of the previous time-step $\Hc_{t-1}$ are first fed into two separate Multi Branch Convolution blocks. Each branch of a Multi Branch Convolution block consists of a different $p\times p$ convolution, $K_p$  followed by a ReLU layer. Afterwards, the output of all the branches are concatenated and fed through a $1\times1$ convolution to generate $\Xb_t$ and $\Hb_{t-1}$, respectively. 

Now we split both $\Xb_t$ and $\Hb_{t-1}$ channelwise into four parts $\left[\Xb_t\right]^1$, $\left[\Xb_t\right]^2$, $\left[\Xb_t\right]^3$, $\left[\Xb_t\right]^4$ and $\left[\Hb_{t-1}\right]^1$, $\left[\Hb_{t-1}\right]^2$, $\left[\Hb_{t-1}\right]^3$, $\left[\Hb_{t-1}\right]^4$. To generate forget gate value $f_t$, the gate responsible for removing some of the information from the previous time step, we run sigmoid on the summation of $\left[\Xb_t\right]^1$, $\left[\Hb_{t-1}\right]^1$ and weight-biased $\C_{t-1}$. Then we move on to calculate the input gate $i_t$, the value that dictates how much new information should be added in the current timestep, in the same manner as we did for forget gate by replacing $\left[\Xb_t\right]^1$ and $\left[\Hb_{t-1}\right]^1$ with $\left[\Xb_t\right]^2$ and $\left[\Hb_{t-1}\right]^2$. For the new information to be generated we run $\tanh$ operation on the summation of $\left[\Xb_t\right]^3$, $\left[\Hb_{t-1}\right]^3$ and $b_c$, learnable parameter we call bias value. New cell value $\C_t$ is calculated by first multiplying $f_t$ with the old cell value $C_{t-1}$ then adding together $i_t$ multiplied the input information values. Now we move on to generate output gate $o_t$, which dictates how information should be let into the new hidden state value $\Hh_t$, by applying sigmoid on the summation of $\left[\Xb_t\right]^4$, $\left[\Hb_{t-1}\right]^4$ and weight-biased new cell value $\C_{t}$. $\Hh_t$ is now calculated simply multiplying $o_t$ with the $\tanh$ed version of the new cell state $C_t$.

The kernel sizes in the multi-branch module are assigned heuristically, and no explicit optimization or empirical search is performed to determine their exact values. Instead, multiple kernels with different receptive fields are employed to enable the network to capture spatial correlations at varying scales. Accordingly, the variation and relative expansion among kernel sizes are considered more important than their absolute values.

\subsection{Fusion Module}
\sys{} maintains 2 encoders: Lightning Encoder encodes lightning observations, and Auxiliary Encoder encodes meteorological condition. Let Lightning Encoder handle $t=-s, -s+1, \ldots, -1$ timesteps of past data and generate the Lightning Cell State, $\C^{light}_{t=-1}$ and the Lightning Hidden State, $\Hc^{light}_{t=-1}$. Likewise, the Auxiliary Cell State, $\C^{aux}_{t=-1}$ and the Auxiliary Hidden State, $\Hc^{aux}_{t=-1}$ get generated by the Auxiliary Encoder. The fusion module fuses the cell ($\C^{light}_{t=-1}$, $\C^{aux}_{t=-1}$) and hidden states ($\Hc^{light}_{t=-1}$, $\Hc^{aux}_{t=-1}$) learned from the \lstm{} of these two encoders, separately. It concatenates the states and passes them through a $1\times1$ convolution (represented as $K_1$ in the following equation) followed by a ReLU layer.
\begin{small}
\begin{align}
    \C^{fused} &= ReLU( K_1 * ( \C^{light}_{t=-1} \mathbin\Vert \C^{aux}_{t=-1} ) )\\
    \Hc^{fused} &= ReLU( K_1 * ( \Hc^{light}_{t=-1} \mathbin\Vert \Hc^{aux}_{t=-1} ) )
\end{align}
\end{small}

\subsection{UpScaler}
The UpScaler module in DeepLight is designed to reconstruct full-resolution lightning prediction frames by incrementally increasing the spatial resolution of the features generated by the decoder. It achieves this by employing a series of transposed convolutional layers, each designed to upscale the feature maps to progressively finer resolutions while preserving spatial correlations learned in the earlier stages. The UpScaler not only reconstructs the high-resolution output but also serves as the final step where the deep representations learned through the dual encoder-decoder architecture are translated into actionable predictions.

\subsection{Hazy Loss}
Traditional loss functions for classification, e.g., Binary Cross Entropy (BCE), only account for the exact matches and mismatches, without considering the spatial or temporal closeness of the predictions made by a model. However, lightning events are inherently uncertain, exhibiting randomness in both space and time. To this end, we propose Hazy Loss, designed to explicitly tackle the spatio-temporal uncertainty intrinsic to lightning occurrence. It enables the model to be trained in a spatio-temporally aware fashion, where the predicted values are based on their distance from the ground truth, rather than by strict correctness alone. The key idea is to introduce a sense of neighbourhood: if a positive prediction is close to a cluster of positive ground truths but not exactly at the place of occurrence, we should penalize the model less. Likewise, when a negative prediction is far away from a ground truth positive cluster, we should not penalize it drastically.

In essence, Hazy Loss allows \sys{} to understand how wrong a prediction is, not just whether it is wrong, which is particularly important for phenomena like lightning that exhibit spatial drift and temporal uncertainty. Lightning occurrences may occur at different locations and times, even under similar observed meteorological conditions. This conceptual understanding motivates the mathematical formulation that follows.

Hazy Loss infuses the spatial and temporal closeness of the predicted values by means of Gaussian blurring~\citep{hummel1987deblurring}. Blurring helps diffuse the closeness information into neighboring cells with gradually diminishing intensity. Let $L_0, L_1, \ldots, L_{h-1}$ represent the ground truth grids corresponding to the prediction horizon of $h$ timesteps. These grids are stacked to form a three-dimensional tensor $L$, which encapsulates the temporal evolution of lightning occurrences across spatial locations. We then apply Gaussian blurring on top of $L$ and normalize it, timestep by timestep, to generate a three-dimensional blurred ground truth tensor $L^{blur}$. Like $L$, we can think of $L^{blur}$ as a tensor where $L^{blur}_0, L^{blur}_1, \ldots, L^{blur}_{h-1}$ are stacked on top of each other. $L^{blur}$ therefore measures how close a cell is to a neighbouring lightning event.

\begin{figure}
    \centering
    \includegraphics[width=0.9\linewidth]{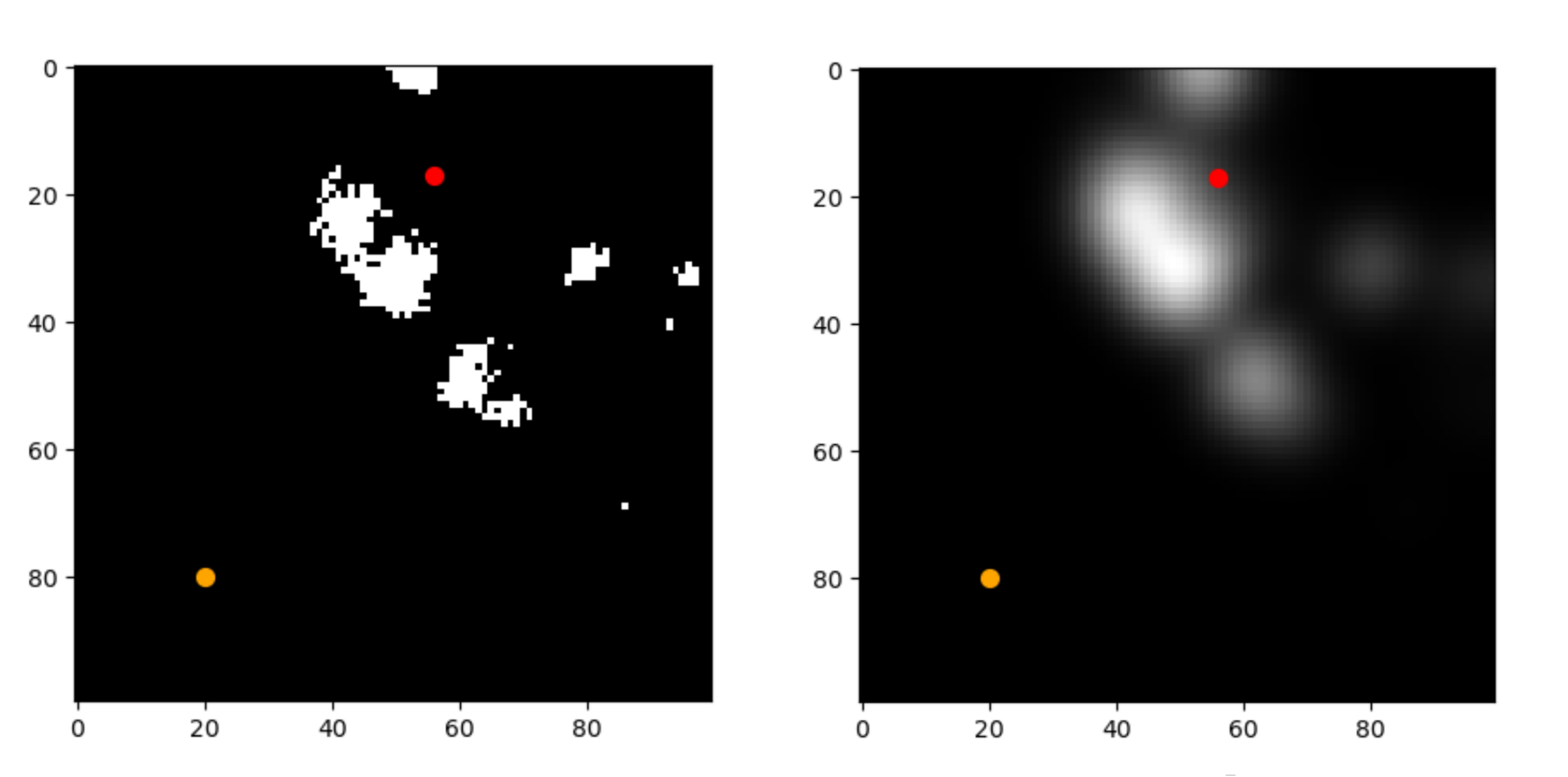}
    \caption{Ground truth grid $L$ (left) and its Gaussian-blurred version $L_{\text{blur}}$ (right) at a particular timestep. White cells indicate lightning occurence and black cells indicate no lightning. \textcolor{red}{Red dot} marks a position close to an actual lightning flash, while the \textcolor{orange}{orange dot} represents a distant location.}
    \label{fig:loss_function_diffusion}
\end{figure}

Figure~\ref{fig:loss_function_diffusion} illustrates how blurring diffuses lightning information into neighboring cells. This blurring allows the loss function to impose a lower penalty when predictions are close to actual lightning flashes (e.g., the red location) and a higher penalty when predictions are farther away (e.g., the orange location).

To calculate the blurred ground truth $L^{blur}$ from $L$, we first generate a 3D Gaussian kernel $K_g$. Note that the size of the kernel is an odd number such that the values at the edge of the kernel are almost zero. Let the shape of the kernel be $\left(s_{x1},s_{x2},s_{x3}\right)$. Then, the input tensor \( L \) is zero-padded before convolution with the Gaussian kernel \( K_g \). The padding is applied symmetrically along each dimension as $\left(\left \lfloor{s_{x1}/2}\right \rfloor, \left \lfloor{s_{x2}/2}\right \rfloor, \left \lfloor{s_{x3}/2}\right \rfloor\right)$. The resulting padded tensor is then convolved with \( K_g \) to generate the blurred ground truth.  

The equations governing this process are provided in Equation~\ref{eq:loss}. Here, \( x_{1d} \), \( x_{2d} \), and \( x_{3d} \) represent the distances from the kernel center to a given kernel cell at \( (x_1, x_2, x_3) \) along their respective dimensions. The parameters \( \sigma_1 \), \( \sigma_2 \), and \( \sigma_3 \) denote the variances along the corresponding axes, controlling the extent of the Gaussian blur.
\begin{small}
\begin{equation}
\begin{aligned}
    &\left[K_{g}\right]_{x_1,x_2,x_3} = \frac{1}{\left(2\pi\right)^\frac{3}{2}\sigma_{1}\sigma_{2}\sigma_{3}}e^{-\frac{x_{1d}^2}{2\sigma_{1}^2}-\frac{x_{2d}^2}{2\sigma_{2}^2}-\frac{x_{3d}^2}{2\sigma_{3}^2}}\\
    &GaussianBlur_{\sigma_1,\sigma_2,\sigma_3}\left(\boldsymbol{L}\right) = K_{g}*Padding\left(\boldsymbol{L}\right)\\
        &\boldsymbol{L'^{blur}} = GaussianBlur_{\sigma_1,\sigma_2,\sigma_3}\left(\boldsymbol{L}\right)\\
        &\boldsymbol{L^{blur}_{t}} = Normalization\left(\boldsymbol{L'^{blur}_{t}}\right)\\
\end{aligned}
\label{eq:loss}
\end{equation}
\end{small}

Given, ground truth $\boldsymbol{L}$ and prediction $\boldsymbol{\hat{L}}$, the Hazy Loss can be computed as follows. 
\begin{small}
\begin{align}
\label{eq:blurring}
    \boldsymbol{P} &= \left(1 - \boldsymbol{L_{blur}}\right) \circ \boldsymbol{\hat{L}} + \boldsymbol{L_{blur}} \circ \left(1 - \boldsymbol{\hat{L}} \right)\\
    \boldsymbol{B} &= - \left(\left(\boldsymbol{L} \circ \log(\boldsymbol{\hat{L}})+(1 - \boldsymbol{L})\circ \log(1 - \boldsymbol{\hat{L}})\right)\right)\\
    Loss_{\text{Hazy}} &= \frac{1}{h\cdot N\cdot N} \left(\boldsymbol{P} \cdot \boldsymbol{B}\right)
\end{align} 
\end{small}
$\boldsymbol{P}$ here is the importance factor of a cell that dictates how much of an impact the BCE loss value of that particular cell will have on the final overall Hazy Loss. After $\boldsymbol{P}$ is calculated, we do the dot product between $\boldsymbol{P}$ the BCE loss vector, $\boldsymbol{B}$ (it is considered as a vector as we compute the BCE loss value of each cell individually) to get a scalar value which is weighted based on the importance factor.

In this context, it is evident that when a cell is in close proximity to a lightning occurrence (spatially or temporally or both), the value of $\boldsymbol{L_{blur}}$ is elevated. Consequently, the weight $P$ for that cell will rely more heavily on the negative prediction ($1-\boldsymbol{\hat{L}}$) than on the positive prediction ($\boldsymbol{\hat{L}}$). If the model prediction $\boldsymbol{\hat{L}}$ is high (the model predicts that there is a high chance of lightning occurrence) the corresponding negative prediction ($1-\boldsymbol{\hat{L}}$) will be low, leading to a reduced weight for the cell due to its comparatively high dependence on the negative prediction. Conversely, a lower prediction value will result in a higher ($1-\boldsymbol{\hat{L}}$) and consequently a higher weight. In contrast, if the cell of interest is located far from a lightning occurrence, $\boldsymbol{L^{blur}}$ will be low, while $1-\boldsymbol{L^{blur}}$ will be high. In this scenario, the weight $P$ will depend more on the positive prediction ($\boldsymbol{\hat{L}}$) than the negative prediction ($1-\boldsymbol{\hat{L}}$), with higher prediction values increasing the weight and lower prediction values reducing it.
 
The Hazy Loss itself is not sufficient to train a model with high prediction accuracy as, by definition, it is a loss based on blurring which removes key information about lightning occurrence. Rather than serving as a standalone loss function, it acts as a complementary aid to train a ML prediction model in a spatio-temporally aware fashion. Thus, we train \sys{} model with the combination of the traditional WBCE and our proposed Hazy Loss, i.e., 
\begin{equation}
    Loss_{\text{Total}} = Loss_{\text{WBCE}} + Loss_{\text{Hazy}}
\end{equation}

Our experiments indicate that the proposed novel loss function significantly enhances the performance not only of DeepLight but also of other machine learning-based lightning prediction models. A comprehensive discussion of these experimental results is provided in Section \ref{subsec:gen_hazy}.

\section{Evaluation}
\label{sec:exp}

In this section, we evaluate DeepLight in experiments. Specifically in the following sections, we show the experiment settings, the comparison of DeepLight with baselines, the effect of Hazy Loss and Multi-Branching based approach, ablation studies, computational efficiency analysis and case studies.

\subsection{Experimental Settings}
\label{sec:settings}
\subsubsection{Dataset}
\label{sec:dataset}
Due to the unavailability of the datasets used in the state-of-the-art lightning prediction models~\citep{lightnet,zhou2022lightnet+,adsnet}, we prepare a new dataset and evaluate \sys{} and the baselines on it. We have made the dataset\footnote{\url{https://doi.org/10.5281/zenodo.15324370}} and code\footnote{\url{https://www.github.com/arifinnasif/DeepLight}} publicly available. The dataset is based on a region of the USA, where lightning is more frequent. The region in contention is centered around Dallas and encompasses certain parts of Texas and Oklahoma. The latitude of the region ranges from \(30.2^\circ N\) to \(35.93^\circ N\) and the longitude ranges from \(93.52^\circ W\) to \(100.3^\circ W\). We divide the region into a grid of \(159 \times 159\) cells with each cell being \(4km \times 4km\). In our experiments, we utilize Lightning Occurrence and Activity data and Cloud Property data from the GOES satellite~\citep{abi-acha, abi-cod, abi-ctp, glm-lcfa}, and Radar Reflectivity data from the NEXRAD radar system~\citep{noaa_radar_reflectivity}, all corresponding to our target region and time frame. The dataset consists of hourly observations collected from April to July for the years 2021, 2022, and 2023. Data from 2021 and 2022 are used for training ($66.66\%$), April and May of 2023 for validation ($16.67\%$), and June and July of 2023 for testing ($16.67\%$). For all experiments, the model uses a lookback window of $s = 6$ hours; that is, the past $s$ hourly frames of lightning observations, radar reflectivity, and cloud properties are used to forecast the next $h$ frames.

\begin{enumerate}
\item \textbf{Satellite Data:} GOES satellite data, available via NOAA's Open Data Dissemination (NODD)\footnote{\url{https://www.noaa.gov/information-technology/open-data-dissemination}}, are accessed using \textbf{goes-2-go} from the \texttt{noaa-goes16} AWS S3 bucket. We obtain cloud top height, cloud top pressure, cloud optical depth, lightning observations, lightning frequency, and flash energy, all in netCDF4 format. Data are interpolated onto a 2D grid using \texttt{scipy.interpolate.griddata}. Table~\ref{tab:param_prod_var} details parameter derivation from various products.

\item \textbf{Radar Data:} NEXRAD S-Band Doppler radars cover the U.S. We use Level 3 Long Range Reflectivity data from the Dallas Lovefield, Texas station (\texttt{NEXRAD:TDAL}), processed via MetPy\footnote{\url{https://www.unidata.ucar.edu/software/metpy/}}. Interpolation follows the same \texttt{scipy.interpolate.griddata} approach. Reflectivity values, ranging from -35dBZ to 65dBZ, are capped at zero for negative values, as per~\citep{noaa_radar_reflectivity}, since they are irrelevant to lightning prediction.
\end{enumerate}

\begin{small}
\setlength\heavyrulewidth{0.20ex}
\begin{table}[htbp]
\ra{1}
\caption{Satellite data overview.}
\begin{tabular}{@{}lclcl@{}} \toprule
Feature & \phantom{abc} & Product & \phantom{abc} & Variable \\
\midrule
    Cloud Top Height      && ABI-L2-ACHA      && `HT'              \\ 
    Cloud Top Pressure    && ABI-L2-CTP       && `PRES'            \\ 
    Cloud Optical Depth   && ABI-L2-COD       && `COD'             \\ 
    Lightning Observation && GLM-L2-LCFA      && `flash\_count'    \\ 
    Flash frequency       && GLM-L2-LCFA      && `flash\_count'    \\ 
    Flash energy          && GLM-L2-LCFA      && `flash\_energy'   \\ 
\toprule
\end{tabular}

\label{tab:param_prod_var}
\end{table}
\end{small}

\subsubsection{Baselines}

We compare \sys{} with the following baselines. 

\begin{itemize}
    \item \textbf{PR92~\citep{price1992simple}.}
    PR92 is a widely used NWP-based lightning parameterization that estimates lightning occurrence as lightning flash rates using a power-law function of convective cloud-top height (CTH): 
    \begin{equation}
        \text{Flash Rate (per minute)}=3.44\times 10^{-5} \text{CTH}^{4.9}
    \end{equation}
    PR92 assumes that deeper convective clouds are associated with stronger updrafts and enhanced charge separation, leading to increased lightning activity. In this work, we implement the continental PR92 formulation using ERA5 reanalysis pressure-level data~\citep{hersbach2020era5}. Cloud-top height is computed as the maximum geopotential height where cloud fraction exceeds a predefined threshold, and the resulting CTH is converted to hourly lightning flash rates using the original PR92 scaling relationship. The outputs are spatially regridded to match the study domain and resolution.

    \item \textbf{Linear Regression~\citep{pearson1901liii}.} Linear Regression is a fundamental statistical method used to model the relationship between a dependent variable and one or more independent variables by fitting a linear equation to observed data. In our implementation, independent linear models predict lightning for each grid cell over a six-hour horizon.
    
    \item \textbf{ST-ResNet~\citep{Zhang_Zheng_Qi_2017}.} ST-ResNet utilizes convolution-based residual networks to effectively capture both nearby and distant spatial dependencies and categorizes temporal attributes into three main aspects: temporal closeness, period, and trend, each modeled by distinct residual networks. It dynamically integrates the outputs from these networks, to enhance predictive performance. In our adaptation, we use a single residual unit, as weekly and monthly trends were not prominent in the lightning data. 
    
    \item \textbf{StepDeep~\citep{stepdeep}.} Originally designed to predict mobility events, StepDeep as a general spatio-temporal framework employs 3D Convolutional Networks for the purpose of learning spatiotemporal features. It integrates a temporal dimension with spatial data through 3-dimensional convolutional kernels, enabling it to effectively predict events in time and space. StepDeep is implemented following the original paper.
    
    \item \textbf{LightNet-O~\citep{lightnet}.} LightNet is a spatio-temporal forecasting model built solely for predicting lightning occurrences. It utilizes data from two different sources, i.e., WRF simulated data, real-world lightning observations, and employs a Dual Encoder-Decoder architecture for predicting lightning occurrences. Its variant, LightNet-O, relies solely on historical lightning observations for its forecasts and is implemented following the original paper\footnote{\url{https://github.com/gyla1993/LightNet}}.
    
    \item \textbf{ADSNet-O~\citep{adsnet}.} ADSNet leverages dual-source data (historical and NWP-based simulated) for lightning prediction, while ADSNet-O focuses specifically on historical lightning observations. Similar to LightNet, ADSNet follows an Encoder-Decoder architecture for lightning predictions. ADSNet-O is modified in our work to support a six-hour prediction horizon\footnote{\url{https://github.com/geolvr/ADSNet}}.
    
    \item \textbf{DeepLight-ViT.}
    To assess the effectiveness of attention mechanisms~\citep{vaswani2017attention} in modeling spatio-temporal lightning occurrences, we introduce a variant of our proposed architecture named DeepLight-ViT. In this baseline, we replace all convolutional components in the original DeepLight architecture with Vision Transformer (ViT) blocks~\citep{dosovitskiy2020image}. This allows us to evaluate whether self-attention can better capture long-range spatial dependencies compared to traditional convolutions. DeepLight-ViT thus serves as a transformer-based counterpart to our CNN-based model, enabling a comparative analysis of attention-driven and convolution-driven representations in the context of lightning forecasting. We replace the convolutional stem with a ViT block in our implementation.
\end{itemize}

Exact settings of all the baselines used can be found in our codebase\footnote{\url{https://github.com/arifinnasif/DeepLight}}. We exclude the \textbf{Hierarchical Spatiotemporal Network (HSTN)~\citep{HSTN}} as a baseline due to the lack of publicly available code and reproducibility concerns.

\subsubsection{Evaluation Metric}
We evaluate the baselines and DeepLight on the following five metrics. Among them, Equitable Threat Score (ETS, for short) is the most important for lightning prediction. ETS adjusts for chance and accounts for the rarity of lightning events, offering a fairer and more reliable performance measurement over rest of the metrics.
\begin{itemize}
    \item POD (Probability of Detection) measures the ratio of correctly predicted lightning events to the total number of observed lightning events.
    \item FAR (False Alarm Rate) measures the ratio of incorrectly predicted lightning events to the total number of predicted lightning events. 
    \item ETS (Equitable Threat Score) measures the accuracy of lightning event predictions while adjusting for hits that could occur purely by chance. Let $N$ denote the total number of grids, and $TP$, $FP$, $FN$, $TN$ denote the True Positives, False Positives, False Negatives and True Negatives, respectively. It is defined as ETS = $\frac{TP-R}{N-TN-R}$, where $R = \frac{(TP+FP)(TP+FN)}{N}$
    \item MicroF1 computes the harmonic mean of precision and recall by globally counting the total true positives, false positives, and false negatives across all prediction instances. 
    \item MacroF1 computes the F1 score separately for both the lightning and no-lightning classes and then averages them. Macro F1 gives equal importance to lightning and non-lightning predictions, making it well-suited for imbalanced settings where rare event detection is critical. 
\end{itemize}

Since both DeepLight and the baseline models produce probabilistic predictions, we apply a predefined threshold to obtain binary lightning/no-lightning labels for metrics based on true and false positives and negatives. We also use a cumulative score metric, defined as follows: for each grid cell $(i,j)$ where a lightning event occurs at time $t$, a true positive is counted if the model predicts lightning in the same cell at any time within the prediction horizon $[t, t+h]$, where $h$ is the horizon length (e.g., 6 hours). This cumulative evaluation strategy, adopted from prior lightning forecasting studies~\citep{lightnet,adsnet}, measures spatial correctness even when the temporal alignment within the horizon is not exact. For example, if a lightning strike occurs at cell $(2,3)$ at 3 PM and the horizon $h$ is 2 hours, a prediction at 4 PM counts as a true positive, while a prediction at 6 PM does not, since it falls outside the horizon window.

\subsubsection{Training Details and Hyperparameters}
The model is implemented using PyTorch 2.3.0 and is trained for $200$ epochs with a learning rate of $0.0001$ on a system configured with 64-bit Windows Server with Intel Xeon Silver 4214R 2.40GHz CPU, 384GB memory, NVIDIA Tesla V100 GPU with 32GB VRAM.  We selected the model with highest validation ETS score.

\begin{figure}
    \centering
    \includegraphics[width=0.75\linewidth]{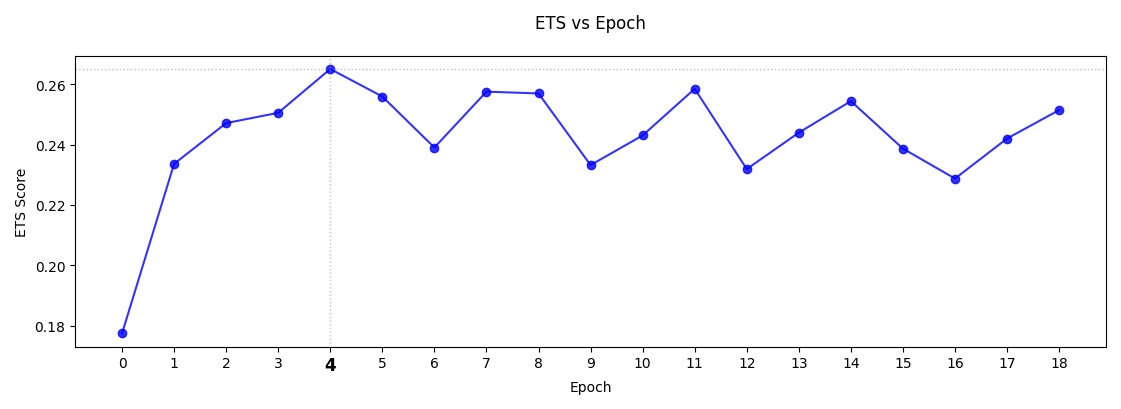}
    \caption{Validation ETS as a function of training epoch during progressive blurring-intensity analysis for Hazy Loss. Peak ETS is observed at epoch 4.}
    \label{fig:whysigma}
\end{figure}

Hazy Loss introduces spatial ($\sigma_1,\sigma_2$) and temporal ($\sigma_3$) Gaussian blur parameters. During preliminary analysis, we apply an exponential decay schedule ($\sigma_1=20\times 0.99^{\text{epoch}}$, $\sigma_2=20\times 0.99^{\text{epoch}}$ and $\sigma_3=1\times 0.99^{\text{epoch}}$) and monitor validation ETS as illustrated in Figure~\ref{fig:whysigma}. The highest ETS is observed at epoch 4, and the corresponding $\sigma_1$, $\sigma_2$ and $\sigma_3$ values are fixed for final model training. Additionally, the positive weight for the WBCE loss function is set to $20$, following values commonly adopted in prior studies~\citep{lightnet,adsnet}.

\setlength\heavyrulewidth{0.20ex}
\begin{table*}[htbp]
\ra{1.3}
\caption{Comparison of \emph{\sys{}} with baselines. Optimal scores are highlighted in bold. $\delta_{ETS}(\%)$ measures the performance improvement of \emph{DeepLight} relative to the model listed in each row. POD (Probability of Detection) and FAR (False Alarm Ratio) are also reported. While a high POD with a low FAR indicates good performance, the ETS (Equitable Threat Score) provides a more balanced metric that accounts for both hits and false alarms, avoiding issues with extreme POD or FAR values~\citep{ebert2008fuzzy, mbizvo2023using, jones2020use}.}

\resizebox{\textwidth}{!}{
\begin{tabular}{@{}lccccccrcccccc@{}} \toprule
\multicolumn{14}{c}{1 hour Cumulative Score}\\
\midrule
\multirow{2}{*}{Method} & \multicolumn{6}{c}{Strict Metric} & \phantom{abc} & \multicolumn{6}{c}{Neighb.-Based Metric} \\
\cmidrule{2-7} \cmidrule{9-14}
& POD & FAR & ETS & MicroF1 & MacroF1 & $\delta_{ETS}(\%)$ && POD & FAR & ETS & MicroF1 & MacroF1 & $\delta_{ETS}(\%)$ \\
\midrule
PR92 & 0.008 & 0.754 & -0.0003 & 0.016 & 0.434 & \textgreater 100 && 0.021 & 0.726 & -0.0001 & 0.038 & 0.435 & \textgreater 100 \\
Linear Regression & 0.120 & 0.988 & 0.001 & 0.021 & 0.481 & \textgreater 100 && 0.235 & 0.970 & 0.007 & 0.053 & 0.475 & \textgreater 100 \\
ST-ResNet & 0.631 & 0.980 & 0.010 & 0.038 & 0.424 & \textgreater 100 && 0.992 & 0.977 & 0.001 & 0.045 & 0.061 & \textgreater 100 \\
StepDeep & 0.810 & 0.753 & \secondbest{0.225} & \secondbest{0.378} & \secondbest{0.682} & 17.8 && 0.913 & 0.564 & \secondbest{0.408} & \secondbest{0.589} & \secondbest{0.789} & 6.9 \\
LightNet-O & 0.846 & 0.777 & 0.206 & 0.352 & 0.668 & 28.6 && 0.935 & 0.604 & 0.375 & 0.556 & 0.771 & 16.3 \\
ADSNet-O & 0.845 & 0.777 & 0.206 & 0.352 & 0.668 & 28.6 && 0.922 & 0.607 & 0.369 & 0.550 & 0.768 & 18.2 \\
\sys{}-ViT & 0.609 & 0.848 & 0.131 & 0.243 & 0.613 & 50.6 && 0.573 & 0.709 & 0.227 & 0.386 & 0.684 & 51.0 \\
\sys{} & 0.758 & 0.703 & \best{0.265} & \best{0.427} & \best{0.708} & x && 0.869 & 0.494 & \best{0.463} & \best{0.640} & \best{0.816} & x \\
\midrule
\multicolumn{14}{c}{3 hours Cumulative Score}\\
\midrule
\multirow{2}{*}{Method} & \multicolumn{6}{c}{Strict Metric} & \phantom{abc} & \multicolumn{6}{c}{Neighb.-Based Metric} \\
\cmidrule{2-7} \cmidrule{9-14}
& POD & FAR & ETS & MicroF1 & MacroF1 & $\delta_{ETS}(\%)$ && POD & FAR & ETS & MicroF1 & MacroF1 & $\delta_{ETS}(\%)$ \\
\midrule
PR92 & 0.020 & 0.648 & -0.0003 & 0.038 & 0.407 & \textgreater 100 && 0.040 & 0.620 & -0.0003 & 0.072 & 0.412 & \textgreater 100 \\
Linear Regression & 0.735 & 0.981 & -0.005 & 0.036 & 0.083 & \textgreater 100 && 0.914 & 0.959 & -0.003 & 0.079 & 0.054 & >100 \\
ST-ResNet & 0.821 & 0.972 & 0.005 & 0.055 & 0.267 & \textgreater 100 && 0.999 & 0.956 & 0.001 & 0.085 & 0.042 & >100 \\
StepDeep & 0.553 & 0.703 & 0.224 & 0.386 & 0.682 & 25.0 && 0.701 & 0.524 & 0.379 & 0.566 & 0.773 & 15.6 \\
LightNet-O & 0.708 & 0.727 & 0.228 & 0.393 & 0.683 & 22.8 && 0.826 & 0.567 & 0.377 & 0.567 & 0.773 & 16.2 \\
ADSNet-O & 0.706 & 0.716 & \secondbest{0.237} & \secondbest{0.404} & \secondbest{0.689} & 18.1 && 0.806 & 0.558 & \secondbest{0.380} & \secondbest{0.570} & \secondbest{0.774} & 15.3 \\
\sys{}-ViT & 0.507 & 0.766 & 0.177 & 0.321 & 0.649 & 36.8 && 0.471 & 0.614 & 0.249 & 0.424 & 0.699 & 43.2 \\
\sys{} & 0.631 & 0.644 & \best{0.280} & \best{0.455} & \best{0.718} & x && 0.741 & 0.462 & \best{0.438} & \best{0.624} & \best{0.805} & x \\
\midrule
\multicolumn{14}{c}{6 hours Cumulative Score}\\
\midrule
\multirow{2}{*}{Method} & \multicolumn{6}{c}{Strict Metric} & \phantom{abc} & \multicolumn{6}{c}{Neighb.-Based Metric} \\
\cmidrule{2-7} \cmidrule{9-14}
& POD & FAR & ETS & MicroF1 & MacroF1 & $\delta_{ETS}(\%)$ && POD & FAR & ETS & MicroF1 & MacroF1 & $\delta_{ETS}(\%)$ \\
\midrule
PR92 & 0.036 & 0.549 & -0.0013 & 0.066 & 0.376 & \textgreater 100 && 0.065 & 0.518 & -0.0017 & 0.115 & 0.385 & \textgreater 100 \\
Linear Regression & 0.975 & 0.959 & -0.001 & 0.080 & 0.052 & \textgreater 100 && 0.997 & 0.927 & 0.000 & 0.137 & 0.070 & \textgreater 100 \\
ST-ResNet & 0.926 & 0.957 & 0.001 & 0.083 & 0.133 & \textgreater 100 && 0.999 & 0.926 & 0.001 & 0.137 & 0.068 & \textgreater 100 \\
StepDeep & 0.352 & 0.668 & 0.184 & 0.341 & 0.656 & 31.0 && 0.485 & 0.485 & 0.311 & 0.499 & 0.737 & 17.7 \\
LightNet-O & 0.514 & 0.693 & 0.212 & 0.384 & 0.673 & 13.7 && 0.651 & 0.531 & \secondbest{0.347} & \secondbest{0.544} & \secondbest{0.757} & 5.5 \\
ADSNet-O & 0.510 & 0.676 & \secondbest{0.222} & \secondbest{0.396} & \secondbest{0.680} & 8.6 && 0.624 & 0.517 & \secondbest{0.347} & \secondbest{0.544} & \secondbest{0.757} & 5.5 \\
\sys{}-ViT & 0.364 & 0.736 & 0.160 & 0.306 & 0.637 & 33.7 && 0.344 & 0.584 & 0.204 & 0.377 & 0.669 & 44.3 \\
\sys{} & 0.456 & 0.616 & \best{0.241} & \best{0.417} & \best{0.694} & x && 0.563 & 0.444 & \best{0.366} & \best{0.559} & \best{0.768} & x \\
\toprule

\end{tabular}
}
\label{tab:evaluation}
\end{table*}

\subsection{Comparison with Baselines}
\label{sec:baselines}
Table~\ref{tab:evaluation} presents a comparative analysis of \sys{}'s performance against the baselines. The evaluation considers three prediction horizons: one hour, three hours and six hours. For each interval, we show their performance using both strict and neighborhood-based metrics to compute the values of true positives, false positives, true negatives, and false negatives. In the strict setting, a lightning event is considered correctly predicted only if it occurs within the exact predicted grid cell. In contrast, the neighborhood-based metric relaxes this condition by also treating predictions as correct if the event falls within any of the eight adjacent grid cells. 

ETS is a more effective metric for lightning prediction, as it accounts for hits, misses, false alarms, and correct rejections, offering a balanced and comprehensive evaluation of model performance~\citep{ebert2008fuzzy, mbizvo2023using, jones2020use}. Among existing learning-based models, LightNet-O~\citep{lightnet} and ADSNet-O~\citep{adsnet} achieve the highest ETS scores, outperforming other baselines such as StepDeep~\citep{stepdeep}, ST-ResNet~\citep{Zhang_Zheng_Qi_2017}, and Linear Regression across all prediction horizons. The PR92 baseline performs poorly across all metrics and horizons, exhibiting extremely low POD and consistently negative or near-zero ETS values. \sys{} significantly outperforms all baselines, achieving the highest ETS across all prediction horizons for both strict and neighborhood-based metrics, with the performance gain being especially pronounced for short-term forecasts. Although LightNet-O~\citep{lightnet} and ADSNet-O~\citep{adsnet} report higher POD values than \sys{}, their substantially higher FAR diminishes their overall effectiveness. By effectively balancing POD and FAR, \sys{} achieves a markedly superior ETS. Furthermore, in terms of both Micro-F1 and Macro-F1 scores, \sys{} consistently outperforms all baseline models, including LightNet-O~\citep{lightnet} and ADSNet-O~\citep{adsnet}, across all prediction horizons and evaluation settings.

Linear Regression is inferior to other models in most metrics because its simple design fails to capture the complex nature of lightning. Linear Regression exhibits a high POD for the 6-hour cumulative forecasts primarily due to temporal aggregation and overprediction. As the accumulation window increases, predicting lightning across large regions increases the likelihood of overlapping with at least one observed event, thereby inflating POD. However, this behavior also leads to an extremely high FAR and near-zero or negative ETS, indicating that the high POD does not reflect meaningful predictive skill. Similarly, while ST-ResNet~\citep{Zhang_Zheng_Qi_2017} exhibits high POD values, it is evident that its FAR is nearly 100\%. This indicates that ST-ResNet~\citep{Zhang_Zheng_Qi_2017} predicts a lightning hit (binary value 1) in all grid cells, suggesting that it is not capable of effectively analyzing lightning data and cannot accurately forecast lightning.

The negative ETS values observed for PR92 arise because the method produces very few true positives while generating a substantial number of false positives, causing the number of correct forecasts to fall below what would be expected from random chance after accounting for chance agreement in the ETS formulation. StepDeep~\citep{stepdeep} outperforms both LightNet-O~\citep{lightnet} and ADSNet-O~\citep{adsnet} for the one hour predictions in all the metrics except POD. However, as the prediction horizon increases, StepDeep's~\citep{stepdeep} performance declines due to its reliance on convolutional filters, which are less effective than recurrent units at capturing long-range temporal dependencies.

The integration of MB-ConvLSTM significantly improves DeepLight’s performance over ConvLSTM-based models~\citep{lightnet, adsnet}. ConvLSTM’s fixed-sized kernel restricts its ability to capture spatial patterns that vary across different lightning events. In contrast, MB-ConvLSTM introduces a multi-branch convolution mechanism, enabling the model to adapt its receptive field to varying spatial extents, thereby improving its ability to predict lightning occurrences across different scales. This ability to handle diverse spatial dependencies, combined with the Hazy Loss function, leads to better performance in terms of ETS across all prediction horizons.

    \noindent\textbf{DeepLight-ViT vs DeepLight:}
As shown in Table~\ref{tab:evaluation}, the attention-based baseline DeepLight-ViT performs worse than the convolution-based DeepLight architecture. DeepLight-ViT struggles to capture the complex spatio-temporal patterns characteristic of lightning prediction. Several factors contributed to this outcome. Firstly, the core strength of DeepLight lies in its simplistic design, which facilitates better generalization. The inclusion of attention mechanisms added significant complexity to the architecture, potentially leading to overfitting on limited training data. This trade-off between model expressiveness and generalization underscores the effectiveness of our minimalist design philosophy. Second, transformer-based attention mechanisms are computationally intensive and typically require large-scale datasets to reach their full potential. Although our dataset is rich, it may lack the scale or diversity necessary to fully leverage transformer capabilities.

\subsection{Hazy Loss as a Generalized Loss Function}
\label{subsec:gen_hazy}
In this study, we evaluate the impact of Hazy Loss on the ETS metric for 1-hour, 3-hour and 6-hour prediction horizons across different models. The performance is compared with the traditional Weighted Binary Cross Entropy (WBCE, for short) Loss. Table~\ref{tab:hazy_ablation} shows the ETS scores of each model under both loss functions, along with the percentage improvement $(\delta)$ achieved using Hazy Loss. The table demonstrates that Hazy Loss consistently improves the ETS scores across all models and prediction horizons compared to WBCE Loss, validating the effectiveness of Hazy Loss in enhancing model performance for lightning prediction tasks. Specifically:
\begin{itemize}
    \item For the \textbf{1-hour} prediction horizon, improvements range from $13.6\%$ to $19.9\%$.
    \item For the \textbf{3-hour} prediction horizon, improvements range from $12.4\%$ to $13.2\%$.
    \item Lastly, for the \textbf{6-hour} prediction horizon, improvements range from $4.5\%$ to $8.1\%$.
\end{itemize}

\setlength\heavyrulewidth{0.20ex}
\begin{table*}[h]
\ra{1.3}
\caption{ Impact of Hazy Loss on Strict ETS Metric Across Different Prediction Horizon on Various Models}
\resizebox{\textwidth}{!}{
\begin{tabular}{@{}lccccccccccc@{}} \toprule
\multirow{2}{*}{Model} & \multicolumn{3}{c}{1 hour} & \phantom{abc} & \multicolumn{3}{c}{3 hours} & \phantom{abc} & \multicolumn{3}{c}{6 hours} \\
\cmidrule{2-4} \cmidrule{6-8} \cmidrule{10-12}
& WBCE Loss only & \makecell{with Hazy Loss} & $\delta(\%)$ && WBCE Loss only & \makecell{with Hazy Loss} & $\delta(\%)$ && WBCE Loss only & \makecell{with Hazy Loss} & $\delta(\%)$ \\
\midrule 
LightNet-O & 0.206 & 0.234 & 13.6\% && 0.228 & 0.258 & 13.2\% && 0.212 & 0.225 & 6.10\% \\
ADSNet-O & 0.206 & 0.247 & 19.9\% && 0.237 & 0.268 & 13.1\% && 0.222 & 0.232 & 4.50\% \\
DeepLight & 0.225 & 0.265 & 17.8\% && 0.249 & 0.280 & 12.4\% && 0.223 & 0.241 & 8.10\% \\

\toprule
\end{tabular}
}
\label{tab:hazy_ablation}
\end{table*}

While Table~\ref{tab:hazy_ablation} reports consistent improvements in ETS obtained by incorporating Hazy Loss, we further evaluate whether these gains are statistically significant across different models and prediction horizons. To this end, we conduct paired two-sided t-tests between model variants trained with and without Hazy Loss.

The tests are performed separately for strict evaluation metrics across 1-hour, 3-hour, and 6-hour prediction horizons. The null hypothesis assumes no performance difference between the two variants trained with and without Hazy Loss. Statistical significance is determined solely based on $p$-values ($\alpha = 0.05$).

Table~\ref{tab:hazy_significance} summarizes the resulting p-values. The results show that Hazy Loss yields statistically significant improvements in most settings, particularly for short- and medium-term forecasts.

\setlength\heavyrulewidth{0.20ex}
\begin{table*}[h]
\ra{1.3}
\caption{Statistical significance of performance gains from Hazy Loss under strict ETS metrics. Paired two-sided \emph{t}-tests are conducted between model variants trained with and without Hazy Loss. Statistical significance is determined at $\alpha = 0.05$.}
\resizebox{\textwidth}{!}{
\begin{tabular}{@{}lcccccccc@{}} \toprule
\multirow{2}{*}{Model} 
& \multicolumn{2}{c}{1 hour} 
& \phantom{ab} 
& \multicolumn{2}{c}{3 hours} 
& \phantom{ab} 
& \multicolumn{2}{c}{6 hours} \\
\cmidrule{2-3} \cmidrule{5-6} \cmidrule{8-9}
& $p$-value & Significant 
&& $p$-value & Significant 
&& $p$-value & Significant \\
\midrule
LightNet-O 
& $2.06\times10^{-24}$ & \cmark 
&& $6.25\times10^{-16}$ & \cmark 
&& $4.29\times10^{-4}$ & \cmark \\
ADSNet-O 
& $1.38\times10^{-4}$ & \cmark 
&& $4.70\times10^{-6}$ & \cmark 
&& $4.68\times10^{-3}$ & \cmark \\
DeepLight 
& $2.67\times10^{-13}$ & \cmark 
&& $1.65\times10^{-4}$ & \cmark 
&& $1.52\times10^{-3}$ & \cmark \\
\toprule
\end{tabular}
}
\label{tab:hazy_significance}
\end{table*}

\subsection{Impact of Multi-Branching}
\label{subsec:effect_of_mb}
\begin{figure}[htbp]
    \centering
    \includegraphics[width=.9\linewidth]{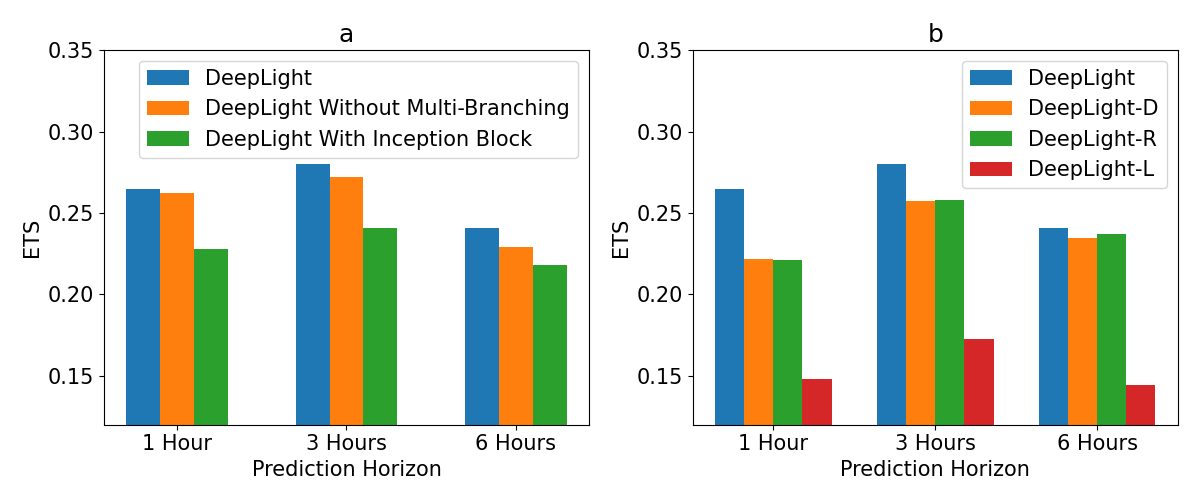}
    \caption{(a) Strict-metric ETS illustrating the effect of Multi-Branching; (b) Strict-metric ETS for different ablated model versions across various prediction horizons.
}
    \label{fig:effect_branch}
\end{figure}

Figure~\ref{fig:effect_branch}a presents the impact of multi-branching on \sys{}'s prediction capability. We evaluate a variant of DeepLight in which all additional branches are removed from both the Convolutional Stem and the LSTM. This variant of DeepLight is unable to capture spatial correlation in multiple extent and thus has poor performance. The ETS score for 1, 3 and 6 hours are $0.262$, $0.272$ and $0.229$, which are $1.13\%$, $2.86\%$ and $4.98\%$ worse than that of \sys{}, respectively. We also evaluate another variant of DeepLight in which every multi-branching block (in both the Convolutional Stem and the LSTM) is replaced with the Inception Block from GoogLeNet~\citep{inception}. This model also under-performs, as the Inception architecture is not well-suited to our specific use case. It has an ETS score of $0.228$, $0.241$ and $0.218$ for 1, 3 and 6 hours, respectively.

\subsection{Ablation Study: Impact of individual features}
\label{sec: ablation}
We perform an ablation study to assess the contribution of individual features by selectively removing them and evaluating its impact on \sys{}'s performance. We evaluate the following three \sys{} variants. 
\begin{itemize}
    \item \emph{\sys{}-D} (L+R) excludes cloud property features (D).
    \item \emph{\sys{}-R} (L+D) excludes radar reflectivity features (R).
    \item \emph{\sys{}-L} (R+D) excludes lightning-related features,  i.e., lightning observations (L) and activities (A). 
    \item \emph{\sys{}} excludes no feature. This is the original model which is trained on all features, i.e. cloud properties, radar reflectivity, and lightning data.
\end{itemize}

\begin{figure*}[t]
    \centering
    \begin{subfigure}{0.55\textwidth}
        \centering
        \includegraphics[width=\textwidth]{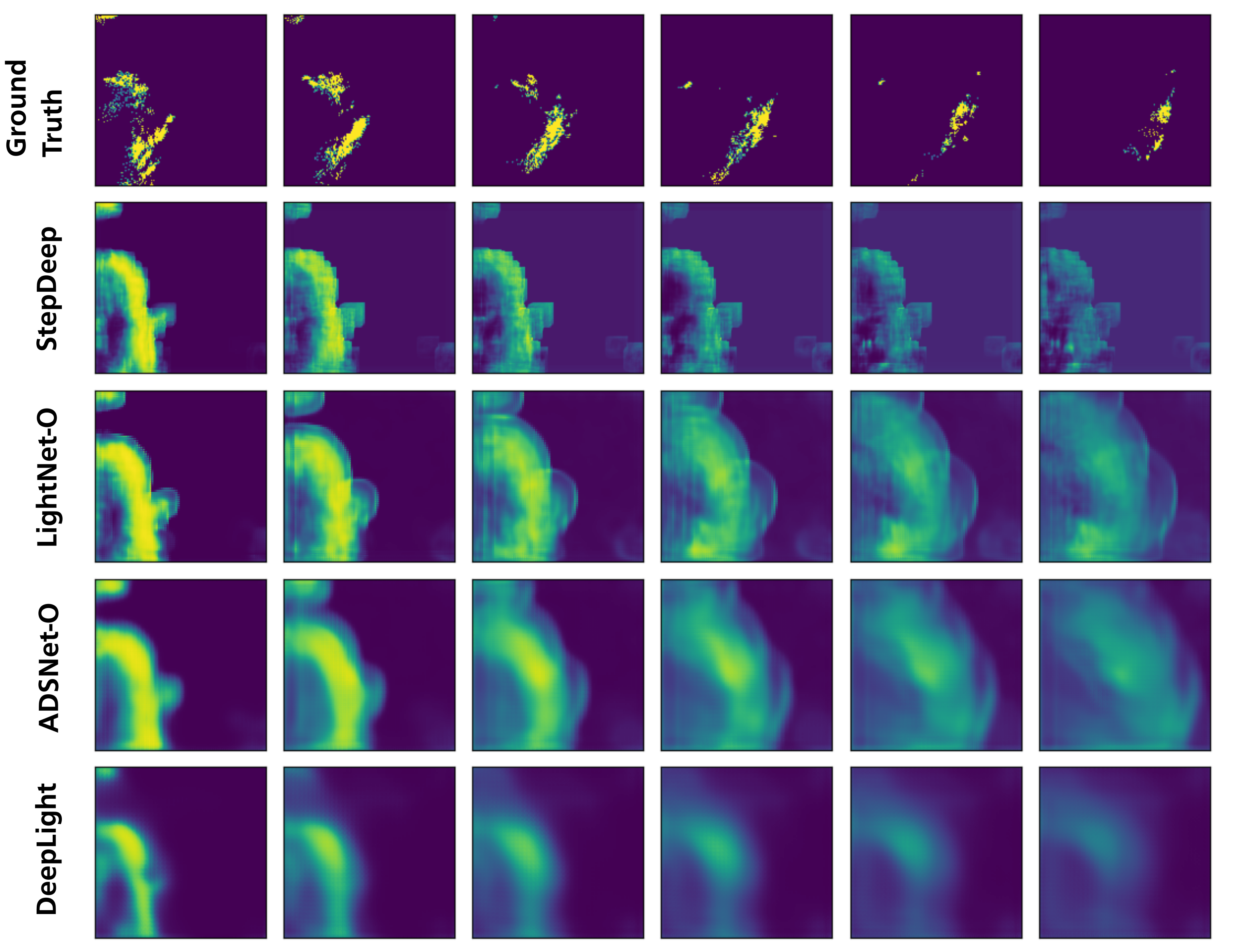}
        \caption{A diminishing thunderstorm}
        \label{fig:casestudy-1}
    \end{subfigure}%
    \begin{subfigure}{0.55\textwidth}
        \centering
        \includegraphics[width=\textwidth]{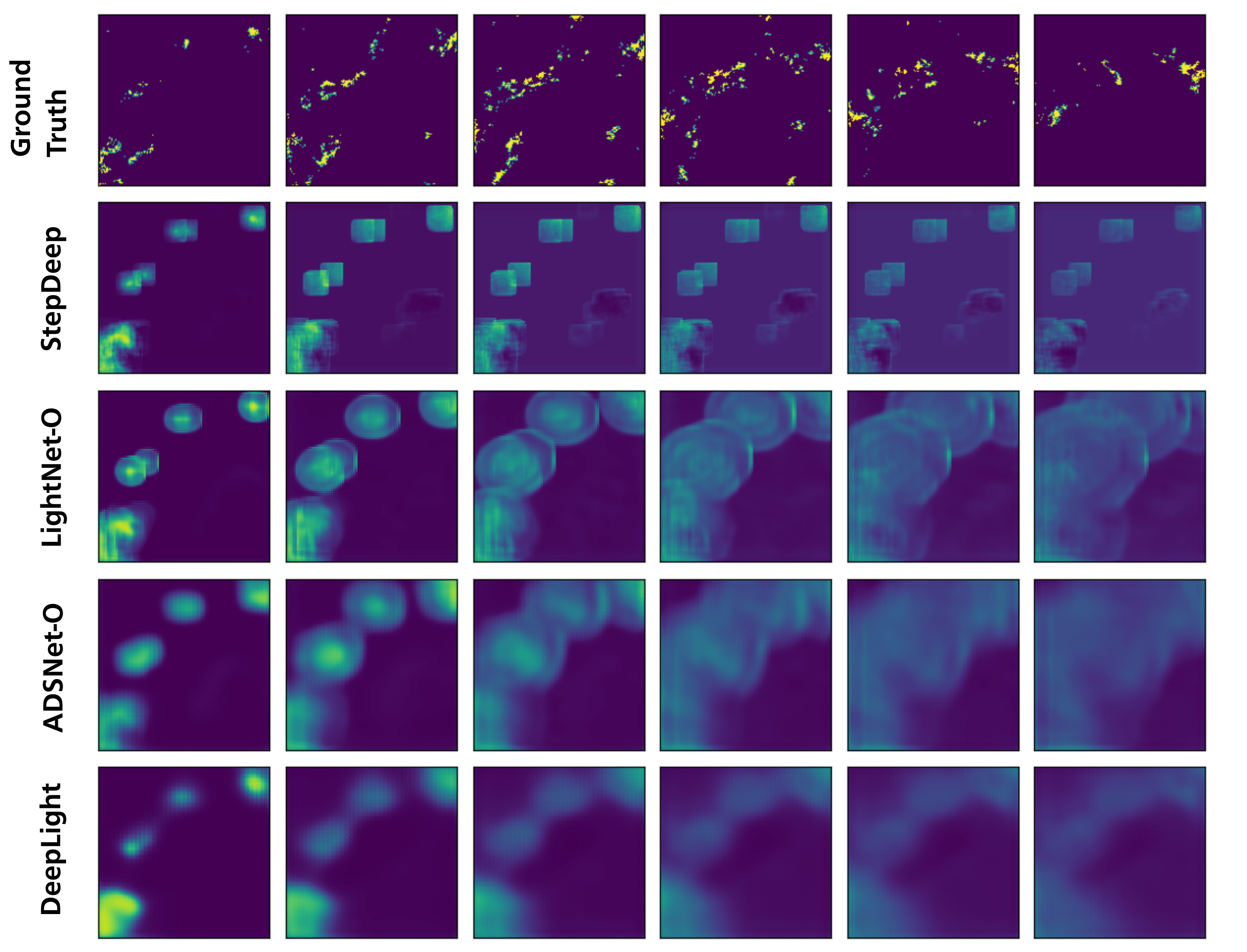}
        \caption{A steady thunderstorm}
        \label{fig:casestudy-2}
    \end{subfigure}
    \caption{Predictions from different models across time steps. Each column corresponds to a specific time step, with the leftmost column representing $t=0$, followed by $t=1$, and so on up to the rightmost column at $t=5$}
\end{figure*}

Figure~\ref{fig:effect_branch}b shows that \sys{} consistently achieves the highest ETS across all prediction intervals, highlighting the importance of all features. Removing lightning features (\sys{}-L) results in the most significant performance drop across all intervals (e.g., a $44.14\%$ drop at 1-hour horizon), highlighting its critical role. Removing cloud properties (\sys{}-D) reduces ETS by $16.25\%$ at 1 hour, $8.17\%$ at 3 hours, and $2.76\%$ at 6 hours, while removing radar reflectivity (\sys{}-R) reduces ETS by $16.52\%$, $7.83\%$, and $1.65\%$, respectively. This shows that cloud properties and radar reflectivity play more significant roles in short-term predictions, improving performance by over $16\%$ in the 1-hour horizon.

\subsection{Computational Efficiency Analysis}
To assess the feasibility of DeepLight for real-time early warning systems, we evaluate its computational efficiency in terms of parameter count, inference latency, memory footprint, and floating-point operations (FLOPs). DeepLight contains $10.85$ million trainable parameters, making it substantially lighter than many attention-based spatio-temporal architectures. Although the training time of our model was almost $18$ hours for $200$ epochs, the inference time of our model is low. We measure inference latency by averaging $50$ forward passes, obtaining an average inference time of $57.2$ ms per sample with a standard deviation of $0.16$ ms. The high training time is not a concern, as the model is trained once and used for inference repeatedly. In terms of memory usage, DeepLight requires a peak CUDA memory allocation of $64.79$ MB during inference. The approximate computational complexity of the model is $1.04 \times 10^{11}$ FLOPs per forward pass.

These results demonstrate that DeepLight achieves a favorable balance between predictive performance and computational cost, supporting its suitability for near real-time lightning early warning applications.

\subsection{Case Study}
\label{sec: case}
The case studies are conducted over the area discussed in Section~\ref{sec:dataset}. This region is characterized by high lightning and thunderstorm activities~\citep{huffines1999lightning}.

\subsubsection{Sample Selection}
Samples with two representative storm events were selected to highlight the models’ behaviors under contrasting scenarios:
\begin{itemize}
\item \textbf{Diminishing Storm:} A thunderstorm that gradually weakens over the six-hour forecast horizon. This scenario allows assessment of models’ ability to capture storm decay and reduce false alarms.
\item \textbf{Stable Clustered Storm:} A storm maintaining relatively stable intensity, with lightning distributed across multiple small clusters. This scenario evaluates the models’ capability to represent fine-scale spatial variability and multi-scale interactions.
\end{itemize}

\subsubsection{Results and Analysis}

\noindent\textit{Diminishing Storm.} Figure~\ref{fig:casestudy-1} illustrates the diminishing storm. In the early frames, all models correctly identify the core lightning region. However, their behaviors diverge as the forecast horizon extends.

StepDeep maintains a nearly static prediction region, only adjusting confidence values. This reflects its limited temporal modeling capacity: the model performs coarse temporal aggregation and lacks mechanisms to capture fine-grained storm decay. ADSNet-O and LightNet-O progressively activate large contiguous regions, indicating spatial overgeneralization. Their convolutional architectures smooth spatial features, reducing sensitivity to local intensity changes and leading to widespread false alarms in later hours.

DeepLight, in contrast, closely follows the observed decay pattern. Its multi-branch convolutional design enables the network to capture both broad storm structures (via large kernels) and localized lightning signatures (via small kernels), providing a theoretically grounded advantage over single-scale convolutional approaches. Additionally, the multi-source fusion mechanism enhances temporal consistency, allowing the network to track weakening storm cells effectively.\\

\noindent\textit{Stable Clustered Storm.} Figure~\ref{fig:casestudy-2} shows a storm with multiple small lightning clusters maintaining stable intensity. StepDeep again produces spatially static predictions, failing to capture the evolving cluster positions. ADSNet-O and LightNet-O expand predictions over the grid, producing widespread but low-intensity activations. This behavior arises from their convolutional architectures, which struggle to represent multi-scale spatial variability and preserve small-scale lightning features.

DeepLight demonstrates superior performance by successfully identifying persistent small clusters and maintaining temporal stability across the six-hour horizon. The multi-branch design allows fine-scale feature extraction while simultaneously capturing broad storm structures. The multi-source representation integrates cloud properties, radar reflectivity, and lightning occurrence, enabling the network to model complex interactions and maintain high spatial precision without overgeneralization. This aligns with the theoretical motivation for using MB-ConvLSTM: multi-branch convolutions allow separate pathways to capture diverse spatiotemporal patterns, which is particularly critical for heterogeneous storms.

\subsubsection{Summary of Insights}
Overall, these case studies confirm the advantages of DeepLight over conventional models:
\begin{itemize}
\item \textbf{Temporal Adaptation:} Multi-source fusion and MB-ConvLSTM enable tracking of storm evolution, including weakening or shifting storm cells.
\item \textbf{Spatial Precision:} Multi-branch convolution captures fine-scale and broad-scale structures, preventing spatial overgeneralization.
\item \textbf{Reduced False Alarms:} Accurate temporal and spatial modeling leads to fewer false predictions, particularly in later forecast hours.
\end{itemize}
The results provide both qualitative and theoretical justification for DeepLight’s architectural innovations, demonstrating its robustness across diverse storm morphologies and validating the improvements observed in Section~\ref{sec:exp}.

\section{Limitation of the Study}
The performance improvement in the 6-hour forecast is relatively modest, ranging from 8\% to 13\%, indicating increased difficulty in long-term lightning prediction. Although DeepLight continues to outperform existing methods, its predictive accuracy decreases as the forecasting horizon extends. This reflects the inherent challenges of modeling the complex, highly dynamic, and chaotic nature of lightning-generating convective systems over longer time scales.

In particular, small uncertainties in storm initiation, evolution, movement, and dissipation can amplify over time, reducing the reliability of long-term predictions. Consequently, the predictive power of short-term observational features—such as lightning history, radar reflectivity, and cloud properties—naturally diminishes for extended forecasting horizons. Addressing these challenges will require future work on more effective modeling of long-term spatio-temporal dependencies and uncertainty.

\vspace{3mm}

\section{Conclusion}
\label{sec:conc}
In this paper, we introduce DeepLight, a deep learning model for lightning prediction that departs from the conventional numerical approach. It features a multi-branch ConvLSTM architecture that extracts spatial correlations from neighborhoods of varying radius. Additionally, DeepLight incorporates a novel neighborhood-aware loss function that penalizes lightning predictions based on their spatio-temporal distance from the ground truth. Our experiments utilize real-world lightning and auxiliary data (i.e., radar reflectivity and cloud properties) derived from GOES satellite and NEXRAD radar. DeepLight outperforms state-of-the-art models in lightning prediction, showing significant ETS improvements: ~30\% for 1-hour, 18–22\% for 3-hour, and 8–13\% for 6-hour horizons. Incorporating Hazy Loss into training further boosts accuracy over the traditional weighted BCE loss, resulting ETS gains of 17.8\%, 12.4\%, and 8.10\% for 1, 3, and 6 hours, respectively. These results highlight the importance of Hazy Loss in enhancing DeepLight’s performance. Furthermore, our computational analysis shows that DeepLight achieves low inference latency and modest memory usage, reinforcing its applicability in real-time operational settings.

Future extensions of DeepLight can focus on expanding both the methodological scope and operational applicability of the model. A promising direction is the integration of additional real-time observational data sources, such as satellite-derived microphysical parameters, environmental soundings, or ground-based electric field measurements, to further enhance predictive robustness. Additionally, extending the framework to probabilistic forecasting and uncertainty quantification may provide more actionable insights for early warning systems, particularly in regions with limited meteorological infrastructure. Finally, deploying DeepLight in real-time operational settings will require optimizing the architecture for faster inference and exploring model compression techniques.




\bibliographystyle{unsrt}
\bibliography{references}

\begin{thebibliography}{10}

\bibitem{cooper2019reducing}
Mary~Ann Cooper and Ronald~L Holle.
\newblock {\em Reducing lightning injuries worldwide}.
\newblock Springer, 2019.

\bibitem{raheem2023techniques}
Babatunte~Dauda Raheem, Emeka Ogbuju, Francisca Oladipo, and Taiwo Abiodun.
\newblock Techniques for lightning prediction: A review.
\newblock {\em Ukrainian Journal of Educational Studies and Information Technology}, 11(4):227--241, 2023.

\bibitem{powers2017weather}
Jordan~G Powers, Joseph~B Klemp, William~C Skamarock, Christopher~A Davis, Jimy Dudhia, David~O Gill, Janice~L Coen, David~J Gochis, Ravan Ahmadov, Steven~E Peckham, et~al.
\newblock The weather research and forecasting model: Overview, system efforts, and future directions.
\newblock {\em Bulletin of the American Meteorological Society}, 98(8):1717--1737, 2017.

\bibitem{price1992simple}
Colin Price and David Rind.
\newblock A simple lightning parameterization for calculating global lightning distributions.
\newblock {\em J. Geophys. Res.}, 97(D9):9919--9933, 1992.

\bibitem{Michalon1999}
N.~Michalon, A.~Nassif, T.~Saouri, J.~F. Royer, and C.~A. Pontikis.
\newblock Contribution to the climatological study of lightning.
\newblock {\em Geophys. Res. Lett.}, 26(20):3097--3100, 1999.

\bibitem{lightnet}
Yangli-ao Geng, Qingyong Li, Tianyang Lin, Lei Jiang, Liangtao Xu, Dong Zheng, Wen Yao, Weitao Lyu, and Yijun Zhang.
\newblock Lightnet: A dual spatiotemporal encoder network model for lightning prediction.
\newblock In {\em KDD}, pages 2439--2447, 2019.

\bibitem{adsnet}
Tianyang Lin, Qingyong Li, Yangli-Ao Geng, Lei Jiang, Liangtao Xu, Dong Zheng, Wen Yao, Weitao Lyu, and Yijun Zhang.
\newblock Attention-based dual-source spatiotemporal neural network for lightning forecast.
\newblock {\em IEEE Access}, 7:158296--158307, 2019.

\bibitem{HSTN}
Yangli-ao Geng, Qingyong Li, Tianyang Lin, Jing Zhang, Liangtao Xu, Wen Yao, Dong Zheng, Weitao Lyu, and Heng Huang.
\newblock A heterogeneous spatiotemporal network for lightning prediction.
\newblock In {\em ICDM}, pages 1034--1039. IEEE, 2020.

\bibitem{zhou2022lightnet+}
Xinyuan Zhou, Yangli-ao Geng, Haomin Yu, Qingyong Li, Liangtao Xu, Wen Yao, Dong Zheng, and Yijun Zhang.
\newblock Lightnet+: A dual-source lightning forecasting network with bi-direction spatiotemporal transformation.
\newblock {\em Appl. Intell.}, 52(10):11147--11159, 2022.

\bibitem{wes-7-1869-2022}
A.~Vemuri, S.~Buckingham, W.~Munters, J.~Helsen, and J.~van Beeck.
\newblock Sensitivity analysis of mesoscale simulations to physics parameterizations over the belgian north sea using weather research and forecasting -- advanced research wrf (wrf-arw).
\newblock {\em Wind Energy Science}, 7(5):1869--1888, 2022.

\bibitem{noaa_radar_reflectivity}
NOAA.
\newblock Radar images reflectivity, Apr 2018.
\newblock Accessed on April 27rd, 2024.

\bibitem{abi-cod}
{GOES-R}.
\newblock Noaa goes-r series advanced baseline imager (abi) level 2 cloud optical depth (cod), 2018.

\bibitem{abi-acha}
{GOES-R}.
\newblock Noaa goes-r series advanced baseline imager (abi) level 2 cloud top height (acha), 2018.

\bibitem{abi-ctp}
{GOES-R}.
\newblock Noaa goes-r series advanced baseline imager (abi) level 2 cloud top pressure (ctp), 2018.

\bibitem{nhess-22-577-2022}
J.~Leinonen, U.~Hamann, U.~Germann, and J.~R. Mecikalski.
\newblock Nowcasting thunderstorm hazards using machine learning: the impact of data sources on performance.
\newblock {\em Natural Hazards and Earth System Sciences}, 22(2):577--597, 2022.

\bibitem{illingworth1985charge}
AJ~Illingworth.
\newblock Charge separation in thunderstorms: Small scale processes.
\newblock {\em J. Geophys. Res.}, 90(D4):6026--6032, 1985.

\bibitem{cintineo2022probsevere}
John~L Cintineo, Michael~J Pavolonis, and Justin~M Sieglaff.
\newblock Probsevere lightningcast: A deep-learning model for satellite-based lightning nowcasting.
\newblock {\em Weather and Forecasting}, 37(7):1239--1257, 2022.

\bibitem{ronneberger2015u}
Olaf Ronneberger, Philipp Fischer, and Thomas Brox.
\newblock U-net: Convolutional networks for biomedical image segmentation.
\newblock In {\em International Conference on Medical image computing and computer-assisted intervention}, pages 234--241. Springer, 2015.

\bibitem{li2022deep}
He~Li, Xuejiao Li, Liangcai Su, Duo Jin, Jianbin Huang, and Deshuang Huang.
\newblock Deep spatio-temporal adaptive 3d convolutional neural networks for traffic flow prediction.
\newblock {\em TIST}, 13(2):1--21, 2022.

\bibitem{liu2023spatio}
Yan Liu, Bin Guo, Jingxiang Meng, Daqing Zhang, and Zhiwen Yu.
\newblock Spatio-temporal memory augmented multi-level attention network for traffic prediction.
\newblock {\em TKDE}, 36(6):2643--2658, 2023.

\bibitem{zhang2018long}
Chaoyun Zhang and Paul Patras.
\newblock Long-term mobile traffic forecasting using deep spatio-temporal neural networks.
\newblock In {\em MobiHoc}, pages 231--240, 2018.

\bibitem{zhang2017deep}
Junbo Zhang, Yu~Zheng, and Dekang Qi.
\newblock Deep spatio-temporal residual networks for citywide crowd flows prediction.
\newblock In {\em AAAI}, volume~31, 2017.

\bibitem{zhang2018predicting}
Junbo Zhang, Yu~Zheng, Dekang Qi, Ruiyuan Li, Xiuwen Yi, and Tianrui Li.
\newblock Predicting citywide crowd flows using deep spatio-temporal residual networks.
\newblock {\em Artificial Intelligence}, 259:147--166, 2018.

\bibitem{tang2022sprnn}
Gaozhong Tang, Bo~Li, Hong-Ning Dai, and Xi~Zheng.
\newblock Sprnn: A spatial--temporal recurrent neural network for crowd flow prediction.
\newblock {\em Information Sciences}, 614:19--34, 2022.

\bibitem{bao2020uncertainty}
Wentao Bao, Qi~Yu, and Yu~Kong.
\newblock Uncertainty-based traffic accident anticipation with spatio-temporal relational learning.
\newblock In {\em ACM-MM}, pages 2682--2690, 2020.

\bibitem{chen2023spatio}
Bingbing Chen and Yong Liao.
\newblock Spatio-temporal deep fusion graph convolutional networks for crime prediction.
\newblock In {\em ICMLSC}, pages 75--81, 2023.

\bibitem{chen2024spatiotemporal}
Xiaoxia Chen, Hanzhong Xia, Min Wu, Yue Hu, and Zhen Wang.
\newblock Spatiotemporal hierarchical transmit neural network for regional-level air-quality prediction.
\newblock {\em Knowledge-Based Systems}, 289:111555, 2024.

\bibitem{huang2021spatio}
Yu~Huang, Josh Jia-Ching Ying, and Vincent~S Tseng.
\newblock Spatio-attention embedded recurrent neural network for air quality prediction.
\newblock {\em Knowledge-Based Systems}, 233:107416, 2021.

\bibitem{lecun1998gradient}
Yann LeCun, L{\'e}on Bottou, Yoshua Bengio, and Patrick Haffner.
\newblock Gradient-based learning applied to document recognition.
\newblock {\em Proceedings of the IEEE}, 86(11):2278--2324, 1998.

\bibitem{elman1990finding}
Jeffrey~L Elman.
\newblock Finding structure in time.
\newblock {\em Cognitive science}, 14(2):179--211, 1990.

\bibitem{Hochreiter}
Sepp Hochreiter and Jürgen Schmidhuber.
\newblock Long short-term memory.
\newblock {\em Neural computation}, 9:1735--80, 12 1997.

\bibitem{convlstm}
Xingjian Shi, Zhourong Chen, Hao Wang, Dit-Yan Yeung, Wai-Kin Wong, and Wang-chun Woo.
\newblock Convolutional lstm network: A machine learning approach for precipitation nowcasting.
\newblock {\em NeurIPS}, 28, 2015.

\bibitem{wang2025application}
Caixia Wang, Xiaoyi Zhang, Hui Yang, Jinyuan Guo, Jia Xu, and Zhuling Sun.
\newblock Application research of convolutional neural network and its optimization in lightning electric field waveform recognition.
\newblock {\em Scientific Reports}, 15(1):1883, 2025.

\bibitem{buyukarikan2022using}
Birkan B{\"u}y{\"u}kar{\i}kan and Erkan {\"U}lker.
\newblock Using convolutional neural network models illumination estimation according to light colors.
\newblock {\em Optik}, 271:170058, 2022.

\bibitem{gong2024spatio}
Yongshun Gong, Tiantian He, Meng Chen, Bin Wang, Liqiang Nie, and Yilong Yin.
\newblock Spatio-temporal enhanced contrastive and contextual learning for weather forecasting.
\newblock {\em IEEE Transactions on Knowledge and Data Engineering}, 36(8):4260--4274, 2024.

\bibitem{stepdeep}
Bilong Shen, Xiaodan Liang, Yufeng Ouyang, Miaofeng Liu, Weimin Zheng, and Kathleen~M Carley.
\newblock Stepdeep: A novel spatial-temporal mobility event prediction framework based on deep neural network.
\newblock In {\em KDD}, pages 724--733, 2018.

\bibitem{wang2023spatial}
Yuchen Wang, Kexin Shi, Chengzhuo Lu, Yuguo Liu, Malu Zhang, and Hong Qu.
\newblock Spatial-temporal self-attention for asynchronous spiking neural networks.
\newblock In {\em IJCAI}, pages 3085--3093, 2023.

\bibitem{lam2023learning}
Remi Lam, Alvaro Sanchez-Gonzalez, Matthew Willson, Peter Wirnsberger, Meire Fortunato, Ferran Alet, Suman Ravuri, Timo Ewalds, Zach Eaton-Rosen, Weihua Hu, et~al.
\newblock Learning skillful medium-range global weather forecasting.
\newblock {\em Science}, 382(6677):1416--1421, 2023.

\bibitem{bi2022pangu}
Kaifeng Bi, Lingxi Xie, Hengheng Zhang, Xin Chen, Xiaotao Gu, and Qi~Tian.
\newblock Pangu-weather: A 3d high-resolution model for fast and accurate global weather forecast.
\newblock {\em arXiv preprint arXiv:2211.02556}, 2022.

\bibitem{bi2023accurate}
Kaifeng Bi, Lingxi Xie, Hengheng Zhang, Xin Chen, Xiaotao Gu, and Qi~Tian.
\newblock Accurate medium-range global weather forecasting with 3d neural networks.
\newblock {\em Nature}, 619(7970):533--538, 2023.

\bibitem{zhang2023long}
Yang Zhang, Lingbo Liu, Xinyu Xiong, Guanbin Li, Guoli Wang, and Liang Lin.
\newblock Long-term wind power forecasting with hierarchical spatial-temporal transformer.
\newblock {\em arXiv preprint arXiv:2305.18724}, 2023.

\bibitem{yao2018deep}
Huaxiu Yao, Fei Wu, Jintao Ke, Xianfeng Tang, Yitian Jia, Siyu Lu, Pinghua Gong, Jieping Ye, and Zhenhui Li.
\newblock Deep multi-view spatial-temporal network for taxi demand prediction.
\newblock In {\em Proceedings of the AAAI conference on artificial intelligence}, volume~32, 2018.

\bibitem{zeiler2010deconvolutional}
Matthew~D Zeiler, Dilip Krishnan, Graham~W Taylor, and Rob Fergus.
\newblock Deconvolutional networks.
\newblock In {\em CVPR}, pages 2528--2535. IEEE, 2010.

\bibitem{inception}
Christian Szegedy, Wei Liu, Yangqing Jia, Pierre Sermanet, Scott Reed, Dragomir Anguelov, Dumitru Erhan, Vincent Vanhoucke, and Andrew Rabinovich.
\newblock Going deeper with convolutions.
\newblock In {\em CVPR}, pages 1--9. IEEE, 2015.

\bibitem{hummel1987deblurring}
Robert~A Hummel, B~Kimia, and Steven~W Zucker.
\newblock Deblurring gaussian blur.
\newblock {\em CVGIP}, 38(1):66--80, 1987.

\bibitem{glm-lcfa}
{GOES-R Algorithm Working Group} and {GOES-R Series Program}.
\newblock Noaa goes-r series geostationary lightning mapper (glm) level 2 lightning detection: Events, groups, and flashes., 2018.

\bibitem{hersbach2020era5}
Hans Hersbach, Bill Bell, Paul Berrisford, Shoji Hirahara, Andr{\'a}s Hor{\'a}nyi, Joaqu{\'\i}n Mu{\~n}oz-Sabater, Julien Nicolas, C{\'e}line Peubey, Raluca Radu, Dinand Schepers, et~al.
\newblock The era5 global reanalysis, 2020.

\bibitem{pearson1901liii}
Karl Pearson.
\newblock Liii. on lines and planes of closest fit to systems of points in space.
\newblock {\em The London, Edinburgh, and Dublin philosophical magazine and journal of science}, 2(11):559--572, 1901.

\bibitem{Zhang_Zheng_Qi_2017}
Junbo Zhang, Yu~Zheng, and Dekang Qi.
\newblock Deep spatio-temporal residual networks for citywide crowd flows prediction.
\newblock In {\em AAAI}, volume~31, 2017.

\bibitem{vaswani2017attention}
Ashish Vaswani, Noam Shazeer, Niki Parmar, Jakob Uszkoreit, Llion Jones, Aidan~N Gomez, {\L}ukasz Kaiser, and Illia Polosukhin.
\newblock Attention is all you need.
\newblock {\em Advances in neural information processing systems}, 30, 2017.

\bibitem{dosovitskiy2020image}
Alexey Dosovitskiy.
\newblock An image is worth 16x16 words: Transformers for image recognition at scale.
\newblock {\em arXiv preprint arXiv:2010.11929}, 2020.

\bibitem{ebert2008fuzzy}
Elizabeth~E Ebert.
\newblock Fuzzy verification of high-resolution gridded forecasts: a review and proposed framework.
\newblock {\em Meteorological Applications: A journal of forecasting, practical applications, training techniques and modelling}, 15(1):51--64, 2008.

\bibitem{mbizvo2023using}
Gashirai~K Mbizvo, Kyle~H Bennett, Colin~R Simpson, Susan~E Duncan, Richard~FM Chin, and Andrew~J Larner.
\newblock Using critical success index or gilbert skill score as composite measures of positive predictive value and sensitivity in diagnostic accuracy studies: Weather forecasting informing epilepsy research.
\newblock {\em Epilepsia}, 64(6):1466--1468, 2023.

\bibitem{jones2020use}
Andrew~S Jones, Allan~A Andales, Jos{\'e}~L Ch{\'a}vez, Cullen McGovern, Garvey~EB Smith, Olaf David, and Steven~J Fletcher.
\newblock Use of predictive weather uncertainties in an irrigation scheduling tool part i: A review of metrics and adjoint methods.
\newblock {\em JAWRA Journal of the American Water Resources Association}, 56(2):187--200, 2020.

\bibitem{huffines1999lightning}
Gary~R Huffines and Richard~E Orville.
\newblock Lightning ground flash density and thunderstorm duration in the continental united states: 1989--96.
\newblock {\em Journal of Applied Meteorology}, 38(7):1013--1019, 1999.

\end{thebibliography}

\end{document}